%% file: uai2025.tex
\documentclass[accepted]{uai2025} % after acceptance, for a revised version; 
\usepackage[american]{babel}
\usepackage{natbib} % has a nice set of citation styles and commands
\usepackage{mathtools} % amsmath with fixes and additions
\usepackage{booktabs} % commands to create good-looking tables
\usepackage{tikz} % nice language for creating drawings and diagrams

\newtheorem{theorem}{Theorem}[section]

\newtheorem{lemma}[theorem]{Lemma}

\newtheorem{assumption}[theorem]{Assumption}
\usepackage[textsize=tiny]{todonotes}
\usepackage{subcaption}
\usepackage{wrapfig}
\usepackage{multicol}
\usepackage{float}
\usepackage{xcolor}
\usepackage{hyperref}
\usepackage{amsfonts}
\usepackage{algorithm}
\usepackage{algpseudocode}

\newcommand{\new}[1]{{\color{black} #1}}
\newcommand{\revise}[1]{{\color{black} #1}}
\newcommand{\algname}{FedSPD}

\title{FedSPD: A Soft-clustering Approach \\ for Personalized Decentralized Federated Learning}

\author[1]{\href{mailto:<ichengl@andrew.cmu.edu>?Subject=FedSPD: A Soft-clustering Approach for Personalized Decentralized Federated Learning}{I-Cheng Lin}{}}
\author[1]{Osman Ya\u{g}an}
\author[1]{Carlee Joe-Wong}
\affil[1]{%
    Department of Electrical \& Computer Engineering\\
    Carnegie Mellon University\\
    Pittsburgh, Pennsylvania, USA
}
  
\begin{document}
\maketitle

\begin{abstract}
  Federated learning has recently gained popularity as a framework for distributed clients to collaboratively train a machine learning model using their local data. While traditional federated learning relies on a central server for model aggregation, recent advancements adopt a decentralized framework, enabling direct model exchange between clients and eliminating the single point of failure. However, existing decentralized frameworks often assume all clients train a shared model. Personalizing each client's model can enhance performance, especially with heterogeneous client data distributions. We propose \textbf{\algname}, an efficient personalized federated learning algorithm for the decentralized setting, and show that it learns accurate models in \textit{low-connectivity} networks. To provide theoretical guarantees on convergence, we introduce a clustering-based framework that enables consensus on models for distinct data clusters while personalizing to unique mixtures of these clusters at different clients. This flexibility, allowing selective model updates based on data distribution, substantially reduces communication costs compared to prior work on personalized federated learning in decentralized settings. Experimental results on real-world datasets show that \textbf{\algname}~outperforms multiple decentralized variants of existing personalized federated learning algorithms in scenarios with \textit{low-connectivity} networks.
\end{abstract}

\input{sections/introduction}

\input{sections/formulation}

\input{sections/algorithms}

\input{sections/math}

\input{sections/simulation_new}

\input{sections/conclusion}

\begin{acknowledgements} % will be removed in pdf for initial submission,
						 % (without ‘accepted’ option in \documentclass)
                         % so you can already fill it to test with the
                         % ‘accepted’ class option
    This work was supported in part by the National Science Foundation (NSF) under Grants CNS-2312761 and CNS-1751075, and by the Office of Naval Research (ONR) under Grant N00014-23-1-2275.
\end{acknowledgements}

% References
\bibliography{uai2025}

\newpage

\onecolumn

\title{Supplementary Material}
\maketitle

\appendix

\input{sections/appendix}

\end{document}

%% file: sections/introduction.tex
\section{Introduction}
\label{Intro}
Federated Learning (FL) is a popular approach for distributed clients to collaboratively learn from their local data. 
The most popular FL algorithm, \textbf{FedAvg} \citep{mcmahan2017communication}, and most of its variants operate within a centralized federated learning (CFL) framework, where a central server coordinates the training process.\footnote{Note that clients in the CFL setting still train their models in a distributed manner; the term "centralized" simply refers to the presence of a central server managing the clients' interactions.} In CFL, each client in a training round independently trains a model on its local data and then sends the model parameters to a central server for aggregation, after which the aggregated model is broadcast back to the clients to begin a new training round. However, communication delays and bottlenecks often arise when a CFL system includes numerous mobile or IoT (Internet-of-Things) clients, hampering CFL’s efficiency. Furthermore, this centralized structure poses risks of attacks and failures due to the single point of failure at the central server \citep{lalitha2018fully}.

\textbf{Decentralized Federated Learning} (DFL) addresses these limitations by adopting a fully decentralized architecture where clients share their locally trained model parameters directly with neighboring clients, eliminating the need for a central server \citep{lalitha2018fully}. This approach can also reduce communication and computational costs \citep{beltran2023decentralized}.
However, most existing DFL methods focus on learning a single global model for all clients, aiming for consensus across clients. Such a global model may under-perform on clients with non-IID (independent and identically distributed) local data, as is commonly the case in federated learning. To address this challenge, we design an efficient \textbf{personalized, decentralized federated learning algorithm} that personalizes models to each client's data distribution without relying on a central server and preserves DFL's communication benefits by limiting the required communication between clients. \revise{We particularly focus on settings where clients are IoT devices using device-to-device communication protocols. Such settings often feature limited network connectivity, communication resources and computation resources, e.g., sensor-based environmental monitoring or vehicles learning personalized models of human driver preferences~\citep{nakanoya2021personalized}.}

Personalization of a shared global model has shown to improve performance in CFL settings \citep{ruan2022fedsoft, marfoq2021federated}. However, extending such personalization methods to DFL poses significant \textbf{technical challenges}. DFL algorithms typically strive for consensus by sharing local models among neighboring clients, which represent only a subset of all clients. Ensuring that all clients can benefit from each other's updates despite limited communication is a key challenge \citep{beltran2023decentralized}. In contrast, learning personalized models requires intentionally maintaining differences in clients' models, particularly for non-IID data. This makes it difficult to distinguish whether model disparities are due to communication issues or differences in local data distributions. We overcome this challenge by \textit{quantifying similarities between client data} using a clustering-based method, allowing the training of distinct models for different data clusters, which are then personalized to each client's unique data mixture.

Prior works that seek to personalize models in DFL settings, including cluster-based methods, are typically straightforward extensions of personalization methods designed for CFL settings, which do not take into account the distinct communication patterns in DFL and thus perform poorly when the client network has poor connectivity. For example, a na\"ive clustering method assigns each client to a single cluster based on its data distribution \citep{ghosh2020efficient}. However, such "hard" clustering assumes identical distributions within the same cluster, which is rarely the case. Instead, we adopt a \textbf{soft clustering} approach, as explored in CFL settings \citep{ruan2022fedsoft, marfoq2021federated}, where each client's data is modeled as an unknown mixture of distributions, and a model is trained for each cluster in this mixture. Existing DFL soft clustering approaches require clients to train models for all clusters in every round \citep{marfoq2021federated}, imposing \textbf{significant training and communication overhead} that scales linearly with the number of clusters. This is particularly problematic in DFL scenarios, where clients often have limited communication and computation capacity \citep{nguyen2021federated}. Therefore, we introduce a training algorithm that (i) learns each client's mixture coefficients, (ii) ensures consensus on models for each cluster, and (iii) unlike prior work, avoids communication resource requirements that scale with the number of clusters. Our \textbf{contributions} are as follows:
\begin{itemize}
    \item We propose \textbf{\algname}, a novel FL algorithm for clients that utilizes soft clustering to train personalized models in a decentralized manner. \textbf{\algname}~allows clients to reach a consensus on cluster-specific models and adapt their cluster mixture estimates over time, while requiring each client to train only \textbf{one} cluster model per training round, significantly reducing communication.
    \item We \textbf{prove the convergence of \algname} in Theorem \ref{thm:4}. This proof adopts a different approach from prior work on soft clustering in DFL, which typically requires clients to train models for every cluster in each round \citep{marfoq2021federated}.
    \item We demonstrate through experiments on real-world datasets that \textbf{\algname~outperforms existing DFL algorithms} (both personalized and non-personalized). In some cases, \textbf{\algname} even approaches the accuracy of centralized algorithms. \revise{Furthermore, we show that \textbf{\algname} is \textbf{particularly effective in low-connectivity networks with computationally constrained clients}}.
\end{itemize}

Following a review of related work in Section~\ref{sec:related}, we present our DFL model in Section~\ref{sec:formulation} and introduce the \textbf{\algname}~algorithm in Section~\ref{sec:algorithms}. We then provide a convergence proof in Section~\ref{sec:math} and demonstrate the algorithm's superior performance in Section~\ref{sec:simulation}, before concluding in Section~\ref{sec:conclusion}.

\section{Related Work}\label{sec:related}
\textbf{Decentralized Federated Learning} has its roots in decentralized optimization \citep{nedic2009distributed, wei2012distributed, zhang2021newton}  and in particular decentralized Stochastic Gradient Descent (SGD) \citep{lian2017can}. Several methods have been explored for decentralized optimization \citep{nedic2009distributed, wu2017decentralized, lu2020computation}, while the convergence analysis of decentralized SGD was first presented by \citet{yuan2016convergence} and \citet{sirb2018decentralized} with delayed information, highlighting decentralized SGD's advantages over centralized methods \citep{lian2017can}. This literature establishes conditions on client connectivity such that all local models will converge to a consensus model \citep{lian2017can}. The effects of client communication topologies in DFL~\citep{lalitha2018fully,warnat2021swarm} have also been studied, and gradient tracking techniques based on push-sum algorithms have been proposed to relax the assumptions on client connectivity needed to show consensus \citep{nedic2014distributed, nedic2016stochastic, assran2019stochastic}.

\textbf{Personalization} in CFL is generally motivated by highly non-IID client data~\citep{mcmahan2017communication, collins2021exploiting}, which can impede convergence and lead to a global model performing poorly at some clients, which may discourage them from participating in the FL process~\citep{huang2020efficiency}.
Common techniques include local finetuning \citep{sim2019personalization}, model interpolation \citep{mansour2020three}, meta-learning \citep{fallah2020personalized}, adding regularization terms \citep{t2020personalized}, and multi-task learning \citep{smith2017federated, yousefi2019multi, li2021ditto}. Clustered FL in particular includes hard clustering, which partitions clients into clusters based on their data's similarity \citep{ghosh2020efficient} and its variations \citep{xie2021multi, briggs2020federated, duan2021fedgroup, mansour2020three}.
In soft clustered FL, one instead assumes that each client's data conforms to a mixture of distributions \citep{marfoq2021federated,ruan2022fedsoft, wu2023personalized}.
Like these prior works, we use models learned for each cluster as guides for a personalized model; unlike them, we add a final personalization step to ensure good performance. We discuss this comparison in more detail in Section~\ref{sec:algorithms}.

Some prior works have considered \textbf{combining personalization and DFL}. \citet{jeong2023personalized} proposed a distillation-based algorithm, while \citet{9993756} proposed a communication-efficient algorithm with model pruning and neighbor selection. \citet{sadiev2022decentralized} prove lower bounds on personalized DFL algorithms' convergence under specific objectives. Unlike these works, we provide theoretical convergence guarantees under more general learning objectives. Some centralized personalization algorithms also include decentralized versions, such as \textbf{FedEM} \citep{marfoq2021federated} and \textbf{IFCA} \citep{ghosh2020efficient}. We experimentally show (Section~\ref{sec:simulation}) that \textbf{\algname}~outperforms both \textbf{FedEM} and \textbf{IFCA}, particularly in low-connectivity settings. Moreover, we \textit{only require each client to train one cluster model at a time}, which leads to significantly smaller computational and communication overhead than \textbf{FedEM}.

\textbf{Comparison with FedSoft.} \textbf{\algname}~was inspired by \textbf{FedSoft} \citep{ruan2022fedsoft}. However, the training methodology is significantly different. \textbf{FedSoft} uses a proximal objective and all client data to update its model in each round, while our \textbf{\algname}~maintains separate models for each cluster and has each client update only one of these models, using only data associated with that cluster, in each round. 
Thus, \textbf{\algname}~avoids bias in gradient updates, which may hamper consensus in decentralized settings. Our theoretical convergence analysis also relaxes the assumptions made by \citet{ruan2022fedsoft} in analyzing \textbf{FedSoft}. We provide a more detailed comparison in Appendix \ref{sec:comp}.

%% file: sections/formulation.tex
\section{Problem Formulation}\label{sec:formulation}

\begin{figure}[t]
\centering
    \includegraphics[width=0.38\textwidth]{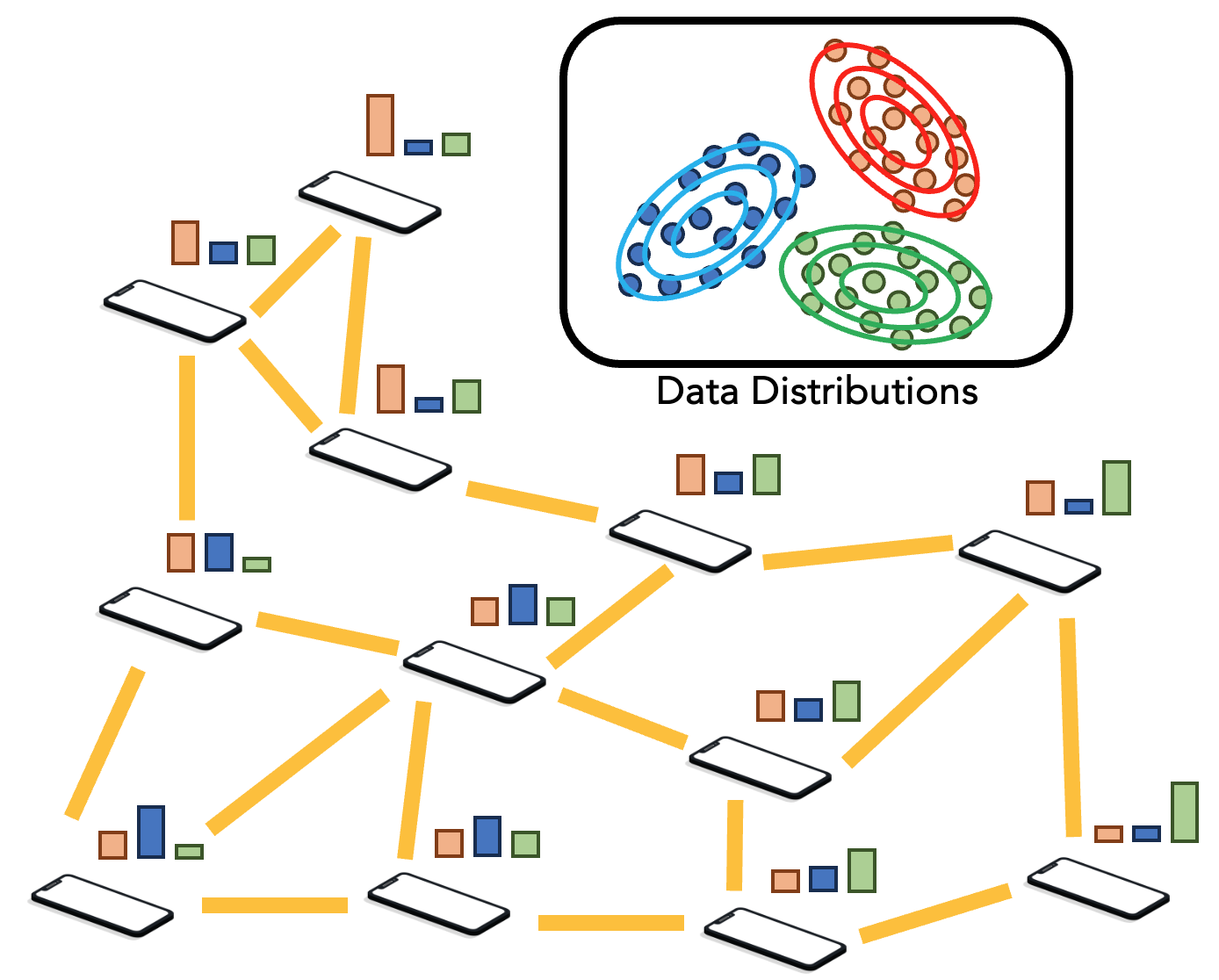}  
\caption{\sl Illustration of the mixture of data distribution at clients in DFL.}
\label{fig:data}
\end{figure}

We illustrate our \textbf{system model} in Figure \ref{fig:data} and summarize our notation in Table~\ref{tab:notation}. We suppose there are $N$ clients that are connected to each other via a graph with adjacency matrix $\mathbf{A}$ and use $\mathcal{N}_i$ to denote the set of client $i$'s neighbors. Each client $i = 1,2,\ldots,N$ has a fixed set $\mathcal{D}_i$ of training data. Clients with a shared edge can directly communicate with each other, e.g., to send model parameters.

\begin{table*}[bht]
  \centering
\begin{tabular}{|p{4.2cm}|p{1.2cm}|p{3.2cm}|p{6.8cm}|}
\hline \textbf{Name} & \textbf{Notation} & \textbf{Domain} & \textbf{Description} \\
\hline Number of Clients / Clusters & $N, S$ & $N, S \in \mathbb{N}$ & Total number of clients / clusters \\
\hline Learning Rate & $\eta_t$ & $\eta_t \in \mathbb{R}, 0<\eta<1$ & Learning rate used in round $t$ \\
\hline Number of Local Updates & $\tau$ & $\tau \in \mathbb{N}$ & Number of local updates in each training round \\
\hline Client Neighbors & $\mathcal{N}_i$ & $\mathcal{N}_i \in \mathcal{P}(N)$ & Indices (in $\left\{1,2,\ldots,N\right\}$) of client $i$'s neighbors \\
\hline Final Model Parameters & $\mathbf{x}_i$ & $\mathbf{x}_i \in \mathbb{R}^{1\times X}$ & Final personalized model parameters of client $i$\\
\hline Final Concatenated Model Parameters & $\mathbf{X}$ & $\mathbf{X} \in \mathbb{R}^{N\times X}$ & Concatenated personalized model parameters\\
\hline Final Phase Epochs & $\tau_{final}$ & $\tau_{final} \in \mathbb{N}$ & Number of epochs for the final phase\\
\hline Local Dataset & $\mathcal{D}_{is}^t$ & $\mathcal{D}_{is}^t \subseteq \mathcal{D}_{i}$, client $i$'s data & Data points at client $i$ associated with cluster $s$ in round $t$ \\
\hline Cluster Selection & $s_{i}^t$ & $s_{i}^t \in \left\{1,2,\ldots,S\right\}$ & Index of cluster that client $i$ trains in round $t$ \\
\hline Portion of Clusters & $u_{is}^t$ & $u_{is}^t \in \mathbb{R}, 0 < u_{is} \leq 1$ & Portion of data for client $i$ of cluster $s$ in round $t$\\
\hline Concatenated Portions of Clusters & $\mathbf{U}(t)$ & $\mathbf{U} \in \mathbb{R}^{N\times S}$ & Concatenated portions of data of all clients in round $t$\\ % \carlee{capitalized or not?} \\
% \hline Cluster Centers & $\mathbf{c}_{is}^t$ & $\mathbf{c}_{is}^t \in \mathbb{R}^{X}$ & Centers of cluster $s$ of client $i$ at time $t$ \\
\hline Average Cluster Centers & $\overline{\mathbf{c}}_{s}^t$ & $\overline{\mathbf{c}}_{s}^t \in \mathbb{R}^{X}$ & Average center of cluster $s$ over clients in round $t$ \\
\hline Concatenated Cluster Centers & $\mathbf{C}_s^t$ & $\mathbf{C}_s^t \in \mathbb{R}^{N\times X}$ & Concatenated centers of cluster $s$ in round $t$ \\
\hline Collection of Cluster Centers & $\mathcal{C}(t)$ & $\mathcal{C}(t) \in \mathbb{R}^{S\times N \times X}$ & $\mathcal{C}(t) = \{ \mathbf{C}_1^t, \mathbf{C}_2^t, ..., \mathbf{C}_S^t \}$ \\
\hline Weight Matrix & $\mathbf{W}_s^t$ & $\mathbf{W}_s^t \in \mathbb{R}^{N\times N}$ & Weight matrix of cluster $s$ in round $t$ \\
\hline Augmented Adjacency Matrix & $\mathbf{A}$ & $\mathbf{A} \in \mathbb{R}^{N\times N}$ & Augmented adjacency matrix with diagonal elements equal to 1 \\
\hline Concatenated Gradients & $\mathbf{G}_s^t$ & $\mathbf{G}_s^t \in \mathbb{R}^{N\times X}$ & Concatenated gradients in round $t$ for cluster $s$, $\mathbf{G}_s^t := [\nabla F_1, ..., \nabla F_N]$\\
\hline
\end{tabular}
\caption{\sl Mathematical notations used in the paper.}
  \label{tab:notation}
\end{table*}

Each data point $d\in \mathcal{D}_i$ on each client $i$ is randomly sampled from one of $S$ unique probability distributions (clusters) denoted as $P_1, P_2, \ldots P_S$, as illustrated in Figure \ref{fig:data}. Consistent with standard clustering methods, we take $S$ as a predetermined hyperparameter \citep{ruan2022fedsoft}. 
Letting $\mathbf{x}$ denote the parameters of a machine learning model, we define the loss function $\ell (\mathbf{x}; \mathcal{D})$ as measuring the sum of the model losses with parameters $\mathbf{x}$ over all points $d$ in a dataset $\mathcal{D}$. Cross-entropy loss, for example, is a typical loss function for classification problems. The \textit{risk} of cluster $s$ can then be written as:
%
%\begin{equation}
    $F_s(\mathbf{x})=\mathbb{E}_{\mathcal{D}\sim P_s}[\ell (\mathbf{x}; \mathcal{D})]$.
%\end{equation}
%
For each client, the risk on a data point $d_{is}$ belonging to cluster $s$ is defined as: $f_{is}(\mathbf{x}, d_{is}) = \ell (\mathbf{x}, d_{is})$. Our goal is for the clients to collectively find the optimal (i.e., risk-minimizing) model parameters for each cluster, which we also call the \textit{cluster centers} and can be written as:
%\begin{equation}
$\mathbf{c}_s^*=argmin_{\mathbf{x}}F_s(\mathbf{x}), \text{ for } s=1, 2, ..., S$.
%\end{equation}
Given the cluster centers and mixture coefficients \( u_{is} \), which represent the proportions of each cluster \( s \) in each client \( i \)'s data, each client can find a personalized model for its local data mixture (Section~\ref{sec:algorithms}). By focusing on common cluster centers, personalized learning can be reframed as achieving consensus on these centers, addressing a key challenge in personalized DFL. However, clients cannot directly determine the cluster centers using their local data \(\mathcal{D}_i\) since it is a \textit{mixture} of clusters, and they do not know which of their data comes from which cluster. In the next section, we present an algorithm for clients to estimate the cluster centers and use them to derive personalized models.

%% file: sections/algorithms.tex
\section{Proposed \algname~Algorithm}\label{sec:algorithms} % 
At each round \( t = 1, 2, \ldots, T \), each client \( i \) maintains two types of parameters: (i) its estimate of the cluster center \(\mathbf{c}_{is}^t\) for each cluster \( s \), and (ii) the cluster to which each data point \( d \in \mathcal{D}_i \) is associated, and the corresponding fraction of its data belonging to each cluster \( s \), denoted by \( u_{is}^t \). In each round \( t \), clients update these parameters based on their local data and information received from their neighbors.

Each round of training consists of \textbf{four steps}: (1) local training, (2) parameter exchange, (3) parameter (i.e., cluster center) update, and (4) data clustering. Following the last training round, we conduct a \textbf{final personalization step}, which involves a local training update to each client's personalized model. Algorithm \ref{alg:CPDFL} formalizes this method.

\textbf{Step 1: Local training} (line 12 in Algorithm \ref{alg:CPDFL}). In round \( t \), each client \( i \) has an estimated portion \( u^t_{is} \) of its data coming from cluster \( s \), where \(\sum_{s=1}^S u^t_{is} = 1\). These values are computed at the end of the previous round (step 4). Client \( i \) then selects cluster \( s \) to update with probability \( u_{is}^t \), ensuring that clients contribute more to clusters where they have more data. By selecting only one cluster per round, \textit{\textbf{\algname}~keeps the training overhead independent of the number of clusters \( S \)}, as each client always trains a single cluster's model. % regardless of \( S \)'s size.

Once a cluster \( s \) is selected, the client performs \( \tau \) SGD updates on its current cluster center estimate \(\mathbf{c}^t_{is}\) using learning rate \(\eta\). Gradients are computed on the risk of the data associated with the selected cluster, \(\mathcal{D}_{i,s}^t\), as \(\nabla_{\mathbf{c}}\ell(\mathbf{c}; d)\), where \( d \) is sampled uniformly at random from \(\mathcal{D}_{i, s}^t\). The dataset \(\mathcal{D}_{i, s}^t\) is formed in the previous round's clustering step, which assigns each data point \( d \in \mathcal{D}_i \) to a cluster. 

\textbf{Step 2: Parameter exchange} (line 18 in Algorithm \ref{alg:CPDFL}). Let \( s_i^t \) be the cluster selected by client \( i \) in round \( t \), meaning that \( \mathbf{c}_{i{s_i^t}}^t \) has been updated. Client \( i \) broadcasts \( s_i^t \) and \( \mathbf{c}_{i{s_i^t}}^t \) to its neighbors $j \in \mathcal{N}_i$. Consequently, each client \( i \) receives the communications \( \{s_j^t, \mathbf{c}_{j{s_j^t}}^t\}_{j \in \mathcal{N}_i} \) from all its neighbors.

\textbf{Step 3: Cluster center updates}

(line 23 in Algorithm \ref{alg:CPDFL}). After receiving the updated cluster centers and indices from its neighbors, each client \( i \) updates the cluster center of the cluster \( s \) it selected to update during this round. The client uses the average of its received cluster centers to update its estimate of \( \mathbf{c}_{is} \):
\begin{equation}
% \small
    \mathbf{c}_{is}^{t+1} = \frac{1}{|j \in \mathcal{N}[i] \cap s_j^t = s|} \sum_{j \in \mathcal{N}[i] \cap s_j^t = s} \mathbf{c}_{js}^t
    \label{eq:cluster_update}
\end{equation}
Here, \(\mathcal{N}[i]\) is the closed neighborhood, including client \( i \) and its neighboring clients, and \( |j \in \mathcal{N}[i] \cap s_j^t = s| \) represents the number of clients \( j \) that both updated cluster \( s \) and belong to \(\mathcal{N}[i]\). 
If no updates for cluster \( s \) are received in round \( t \), i.e., none of the neighbors selected it, the estimated cluster center remains unchanged: \(\mathbf{c}_{is}^{t+1} = \mathbf{c}_{is}^{t}\). This update rule can be expressed in matrix form as \(\mathbf{C}_{s}^{t+1} = \mathbf{W}_{s}^t \mathbf{C}_{s}^{t}\), where \(\mathbf{W}_{s}^t\) is the weight matrix for cluster \( s \) at time \( t \), and \(\mathbf{C}_{s}^{t} = [\mathbf{c}_{1s}^{t}, \dots, \mathbf{c}_{Ns}^{t}]\) contains the concatenated cluster centers.

\textbf{Step 4: Data clustering}
(line 29 in Algorithm \ref{alg:CPDFL}). After updating the cluster centers, each client \( i \) associates its data points \( d \in \mathcal{D}_i \) with a cluster. It calculates the loss \(\ell(\mathbf{c}_{is}^{t+1}, d)\) for each cluster \( s \) and assigns data point $d$ to the cluster with the lowest loss. Using these new associations, \( u_{is}^{t+1} \), the fraction of data points linked to cluster \( s \), is computed. This step enables \textbf{\algname} to adapt the mixture coefficients as cluster center estimates evolve. The process then moves to the next round, \( t+1 \), starting again with local training.

\textbf{Final Step: Personalization} %\label{sec:personalization}
(line 37 in Algorithm \ref{alg:CPDFL}). After \( T \) rounds, each client $i$ computes a personalized model as a weighted sum of its cluster centers:
\begin{equation}
%\small
    \mathbf{x}_i = \sum_{s=1}^S u_{i, s}^T \mathbf{c}_{i, s}^T
    \label{eq:xi}
\end{equation}
\citet{marfoq2021federated} show that Eq.~\eqref{eq:xi} provides the optimal personalized model for client \( i \) when the loss function \( \ell \) is convex. However, since most practical loss functions, such as cross-entropy for neural networks, are not convex, this aggregated model may perform poorly in practice. Thus, each client runs a few additional local training iterations, starting from \(\mathbf{x}_i\) (Eq.~\eqref{eq:xi}), using its entire local dataset \(\mathcal{D}_i\).

\textbf{Comparison to prior soft clustering algorithms.} \citet{marfoq2021federated} and \citet{ruan2022fedsoft} use soft clustering to learn cluster centers and personalized models without this final personalization step, directly learning personalized models in each iteration, with a central server estimating the cluster centers. In DFL, achieving consensus on cluster models is difficult due to the extensive parameter exchanges needed for model propagation, particularly when clients have few neighbors. \citet{marfoq2021federated} propose a decentralized algorithm that sets the personalized model as a weighted sum of the cluster centers \textit{at each round's end}, which can be sub-optimal for non-convex loss functions. Such a framework can lead to overfitting in DFL, as clients have low connectivity and thus cannot rely on receiving many other clients' updates in each training round. Adding another final personalization step, as we use in \textbf{\algname}, may exacerbate this overfitting, as cluster center gradients already incorporate personalized models. In Section~\ref{sec:simulation}, we demonstrate that \textit{\textbf{\algname}~outperforms \citet{marfoq2021federated}'s \textbf{FedEM} algorithm}, which also requires each client to train all models per round, incurring significantly more computation and communication than \textbf{\algname}.
\begin{algorithm}[h!]
\caption{Our Proposed \textbf{\algname} Algorithms}
\label{alg:CPDFL}
% \begin{multicols}{2}
\begin{algorithmic}[1]
\Procedure {FedSPD}{$\eta$, $\tau$, $S$, $T$, $\mathbf{W}_s^t$}
\For {$t = 1, 2, ..., T\tau$}
\State \textproc{LocalUpdate}($\mathcal{C}(t)$)
\If {$t \: mod \: \tau =0 $}
\State \textproc{ParameterExchange}($\mathcal{C}(t)$, $\mathbf{A}$)
\State \textproc{ParameterUpdate}($\mathcal{C}(t)$, $\mathbf{A}$)
\State \textproc{DataClustering}($\mathcal{C}(t)$, $\mathbf{A}$)
\EndIf
\EndFor
\State \textproc{FinalPhase}($\mathcal{C}(t)$, $\mathbf{u}(t)$)
\EndProcedure
\Procedure {LocalUpdate}{$\mathcal{C}(t)$}
\For {$i = 1, 2, ..., N$}
\State Client $i$ selects cluster $s_i^t$ to update
\State $\mathbf{c}_{s_i^t}^{t+1} = \mathbf{c}_{s_i^t}^{t} - \eta_t \nabla f_{is} (\mathbf{c}_{s_i}^{t})$
\EndFor
\EndProcedure
%\\
\Procedure {ParameterExchange}{$\mathcal{C}(t)$, $\mathbf{A}$}
\For {$i = 1, 2, ..., N$}
\State For each client $i$, exchange the updated parameter $\mathbf{c}_{is}$ and the selected cluster $s$ with client $j \in \mathcal{N}_i$
\EndFor
\EndProcedure
%\\
\Procedure {ParameterUpdate}{$\mathcal{C}(t)$, $\mathbf{A}$}
\State Construct $\mathbf{W}_{s}^t$ for each cluster $s$. If client $i$ is not selected to update cluster $s$, the row $i$ and column $i$ will only have diagonal element equal to 1, else equal to 0 , meaning the model parameter of the cluster that a user has not selected to update will remain the same as it was in the previous epoch.
\For {$s = 1, 2, ..., S$}
\State $\mathbf{C}_{s}^{t+1} = \mathbf{W}_{s}^t \mathbf{C}_{s}^{t+1}$
\EndFor
\EndProcedure
%\\
\Procedure {DataClustering}{$\mathcal{C}(t)$, $\mathbf{A}$}
\For {$i = 1, 2, ..., N$}
\For {$d_k \in \mathcal{D}_i$}
\State Label data $d_k$ with the least loss of all the model parameters among all clusters.
\EndFor
\State For $s=1,...,S$ update $u_{i, s}^t$ for client $i$
\EndFor
\EndProcedure
%\\
\Procedure {FinalPhase}{$\mathcal{C}(t)$, $u^t$}
\For {$i = 1, 2, ..., N$}
\State $\mathbf{X}_{i} = \sum_{s=1}^S u^t_{i, s} \mathbf{C}_s^t(i, :)$
\EndFor
\For {$t = 1, 2, ..., \tau_{final}$}
\State \textproc{LocalUpdate}($\mathbf{X}$) \algorithmiccomment{Run gradient descent using all data of the client for the aggregated training.}
\EndFor
\EndProcedure
\end{algorithmic}
% \end{multicols}
\end{algorithm}

%% file: sections/math.tex
\section{Convergence Analysis}\label{sec:math}
We prove that \textbf{\algname}~converges in Theorem \ref{thm:4}. We first outline our technical assumptions and then present our main results. All proof details can be found in Appendix \ref{sec:proof}.

\textbf{Assumptions.} Our analysis relies on the following  assumptions on the risk function and gradient estimates, which are common in the literature~\citep{marfoq2021federated, ghosh2020efficient, koloskova2020unified} and weaker than those of~\citet{ruan2022fedsoft}.

\begin{assumption}\label{as:1}
(Strong convexity and smoothness) The risk function $F_s$ for each cluster $s$ is $L$-smooth and $\mu$-strongly convex. That is, for some $L>0$ and $\mu \geq 0$:
\begin{equation}
%\small
\begin{aligned}
    & \| \nabla F_s(\mathbf{x}) - \nabla F_s(\mathbf{y}) \| \leq L \|\mathbf{x} - \mathbf{y} \| ; \\
    & \nabla F_s (\mathbf{x})^T(\mathbf{y}-\mathbf{x}) + \frac{\mu}{2} \|\mathbf{y} - \mathbf{x}\|^2 \leq F_s(\mathbf{y}) - F_s(\mathbf{x})
\end{aligned}
\end{equation} % \carlee{where is $\mu$?}
\end{assumption}

\begin{assumption}\label{as:2}
(Bounded risk function) The risk function $F_s$ for each cluster $s$ is lower-bounded by some $F_{inf} > 0$, i.e., 
%\begin{equation}
    $F_s(\mathbf{x}) \geq F_{inf}$.
%\end{equation}
%for some $F_{inf} > 0$.
\end{assumption}

\begin{assumption}\label{as:3}
(Unbiased gradient estimation) The gradient is unbiased, i.e.,
%\begin{equation}
    $\mathbb{E}[\nabla f_{is}(\mathbf{x})]=\nabla F_s(\mathbf{x})$.
%\end{equation}
\end{assumption}

\begin{assumption}\label{as:4}
(Bounded gradient) We have
%\begin{equation}
    $\mathbb{E} \| \nabla f_{is}(\mathbf{x})\|^2 \leq \sigma^2$
%\end{equation}
for some $\sigma^2 > 0$.
\end{assumption}

\begin{assumption}\label{as:5}
(Bounded variance of gradient estimation) The variance of the estimated gradient is bounded:
\begin{equation}
% \small
    \mathbb{E} \| \nabla f_{is}(\mathbf{x}) - \nabla F_s(\mathbf{x}) \|^2 \leq v^2 \textnormal{, for some } v^2 > 0.
\end{equation}
%for some $v^2 > 0$
\end{assumption}

\begin{assumption}\label{as:6}
(Bounded cluster error) Following \citet{ruan2022fedsoft, ghosh2020efficient}, \revise{during all training steps $t$}, all estimated cluster centers have bounded distance to the optimal centers. That is, for some $\delta > 0$: % \textcolor{blue}{C: for some $\delta > 0$}:
\begin{equation}
%\small
    \| \mathbf{c}_{is}^t - \mathbf{c}_{s} ^* \| \leq (0.5 - \alpha_0) \sqrt{\frac{\mu}{L}} \delta, \forall s \in 1, 2, ..., S
\end{equation}
where $0 < \alpha _0 \leq 0.5$. % \carlee{for sufficiently large $t$?}. Without loss of generality, we also assume for all $s$, $\| \mathbf{c}_s^{\star}\| \leq 1$.
\end{assumption}
Note that this assumption will always hold for some value of $\delta$; however, a larger $\delta$, and thus larger cluster error, will also lead to slower convergence.

We finally follow \citet{koloskova2020unified} in assuming that clients communicate sufficiently for consensus:

\begin{assumption}\label{as:7}
(Expected consensus rate) Define $\mathbf{C_s}$ as the concatenated model parameter matrix of cluster $s$. Then for some constant $p \in (0, 1]$ and integer $\beta \geq 1$, for all non-negative integers $l \leq \frac{T}{\beta}$ we have: 
\begin{equation}
%\small
    \mathbb{E} \left\| \mathbf{C}_s \prod_{t =l \beta}^{(l+1)\beta -1}\mathbf{W}_s^t - \overline{\mathbf{C}_s}\right\|_F^2 \leq (1-p) \| \mathbf{C}_s - \overline{\mathbf{C}_s} \|_F^2
\end{equation}
where $\bar{\mathbf{C}_s} := \underbrace{[\bar{\mathbf{c}_s}, ..., \bar{\mathbf{c}_s}]}_{\text{total }N\text{ terms.}}$ is the matrix with every column equal to the average of the model parameters.
\end{assumption}

For simplicity, we further assume that all clients have the same amount of data (i.e., $\mathcal{D}_i$ has the same number of data points for all clients $i$) and that the number of local updates $\tau = 1$ in the remainder of this section. These can be easily relaxed if needed.

\textbf{Results.} Without loss of generality, we present our results for a specific cluster \( i \), where \( i = 1, \dots, S \). Since the convergence proof is identical for each of the \( S \) clusters, we omit the cluster index for clarity. Let \( n \) be the number of clients chosen to update the selected cluster. If the total data across clients is roughly uniform for each cluster, then \( n \approx \frac{N}{S} \). We begin by bounding the distance of the average cluster center to its optimal center:
\begin{theorem} \label{thm:1}
    (Descent lemma) The distance $\mathbb{E}\left\|\overline{\mathbf{c}}^{(t+1)}-\mathbf{c}^{\star}\right\|^2$ between the average cluster center and its optimum $\mathbf{c}^\star$ satisfies the bound (\ref{thm1eq}) with proper choice of learning rate $\eta_t$:
    \begin{equation}
    % \scriptsize
%    \small
    \begin{aligned}
      &\leq  \frac{\eta_t (L+\mu)}{n}\sum_{i=1}^{n_1}\left\|\overline{\mathbf{c}}^{(t)}-\mathbf{c}_i^{(t)}\right\|^2 + \frac{18L^2\epsilon_N^2 \eta_t^2}{n^2} + v^2\eta_t^2 \\
    & + \left(1- \eta_t \mu + \frac{\eta_t \mu\epsilon_N}{n}\right)\left\|\overline{\mathbf{c}}^{(t)}-\mathbf{c}^{\star}\right\|^2 + \frac{2\epsilon_N (S-1) v^2 \eta_t^2}{n^2} \\
    &  + \left(\frac{4\eta_t^2(n-\epsilon_N)^2 L}{n^2} + 2\eta_t\frac{1 - \epsilon_N}{n}\right) \left(f\left(\overline{\mathbf{c}}^{(t)}\right)-f\left(\mathbf{c}^{\star}\right)\right)
    \end{aligned}
    \label{thm1eq}
    \end{equation}
    Here $\epsilon_N$ is the bound of the expected number of clients using the wrong data in Lemma \ref{lm:ep}.
\end{theorem}
We then derive an expression for the cluster centers estimated by individual clients.
\begin{theorem} \label{thm:2}
    (Update rule) % The model parameters
    Clients' estimated centers of the cluster after time $t$ can be written as: 
    \begin{equation}
%    \small
    \begin{aligned}
    \mathbf{C}^{t} & = \mathbf{C}^{l\beta}\prod_{m=l\beta}^{t-1}\mathbf{W}^{m}-\sum_{m=l\beta}^{t-1}\left(\eta_t\mathbf{G}^{m} \prod_{r=t-1}^{m}\mathbf{W}^{r}\right)
    \end{aligned}
    \end{equation}
\end{theorem}
Here $l \in \mathbb{N}$ and $\beta$ is the constant in Assumption \ref{as:7}. Given this expression, we can relate the clients' cluster center estimates to their average, showing that they eventually reach a near-consensus:
\begin{theorem} \label{thm:3}
    (Consensus distance) Define $\mathbf{E}_t = \frac{1}{N}  \sum_{i=1}^N \mathbb{E} \| \mathbf{c}_i^{(t)}- \overline{\mathbf{c}}^{(t)}\|^2$, the expected squared %\carlee{(squared)} 
    distance of the model parameters of client $i$ to the average model parameter. It is upper-bounded by
\begin{equation*}
%\small
\begin{aligned}
\left(1-\frac{p}{2}\right) \mathbf{E}_{m\beta} + & \sum_{j=m\beta}^{t-1}  \bigg(\frac{p\mathbf{E}_j}{16\beta} + \frac{18\beta n \sigma^2 + nv^2p}{Np} \eta_j^2 \\
& {} \quad + \left(\frac{36Ln\beta\eta_t^2}{pN}\right) (f(\overline{\mathbf{c}^{{j}}})-f(\mathbf{c}^{\star})) \bigg)
\end{aligned}    
\end{equation*}
Here $p$ is the constant defined in Assumption \ref{as:7}. 
\end{theorem}

\begin{theorem} \label{thm:4}
(Cluster convergence rate)
For given target accuracy $\epsilon$, there exists a constant learning rate for which $\epsilon$ accuracy can be reached after $T$ iterations: 
\begin{equation}
%\small
\begin{aligned}
    \left[1+\left(\frac{n-\epsilon_N}{n}\right) \eta L \right] & \sum_{t=0}^T\frac{r_t}{R_T}(\mathbb{E}f(\overline{\mathbf{c}}^{t})-f({\mathbf{c}^{\star}})) \\
    & + \mu \mathbb{E}\| \overline{\mathbf{c}}^{(T+1)} - \mathbf{c}^{\star}\|^2 \leq \epsilon
\end{aligned}
\end{equation}
Here $r_t$ is a sequence of positive weights defined in Lemma~\ref{lm:cr} in Appendix~\ref{sec:thm4-proof} %\textcolor{blue}{C: $w_t$ doesn't appear in (9), so I don't think we need it here?} 
and $R_T = \sum_{t=1}^T r_t$. Rearranging, we find that the number of required iterations $T$ to reach an error $\epsilon$ is of the order:
\begin{equation*}
%\small
\begin{aligned}
    \tilde{\mathcal{O}} \Bigg( \frac{\sqrt{L+\mu}}{\sqrt{\epsilon} \mu}\bigg(\sigma & + \frac{\sqrt{n}}{\sqrt{N}}v\bigg) + \frac{L\beta n^{\frac{3}{2}}}{\mu p \sqrt{N}(n-\epsilon_N)}\ln\left(\frac{1}{\epsilon}\right) \\
    & + v^2 \frac{n^2 + L^2 \epsilon_N^2 + \epsilon_N (S-1)}{\mu n^2 \epsilon} \Bigg). \\
\end{aligned}
\end{equation*}
\end{theorem}
\new{The convergence rate, asymptotically requiring $O(1/\sqrt{\epsilon})$ training rounds to reach an error $\epsilon$, aligns with previous works on DFL without personalization \citep{koloskova2020unified}%\carlee{cite these works}
, leading us to conjecture that \textbf{\algname}~will converge well. We note that the network connectivity appears in this bound through the constant $p \in (0, 1]$ (Assumption \ref{as:7}), where higher connectivity indicates a larger $p$. However, the second term in the convergence rate that involves $p$ is not the dominant term. Thus, as long as the network is connected, we expect that the effect of network connectivity on convergence will be relatively minor. Our simulation results in Section \ref{sec:sim_nc} support this observation.}

%% file: sections/simulation_new.tex
\section{Simulation Results}\label{sec:simulation}

In this section, we evaluate the performance of our proposed \textbf{\algname}~algorithm and compare it with existing methods. We also analyze how different network connectivity and topology influence \textbf{\algname}'s performance.

\textbf{Datasets and models.} Unless specified, we use \( N = 100 \) clients for all experiments on hand-written character recognition (MNIST and EMNIST datasets \citep{cohen2017emnist}) and \( N = 25 \) clients for all experiments on image classification (CIFAR-10 and CIFAR-100 datasets \citep{krizhevsky2009learning}). We use a CNN (convolutional neural network) model for each client, following the settings of \cite{ruan2022fedsoft} with data from a mixture of \( S = 2 \) distributions, \(\mathcal{D}_A\) and \(\mathcal{D}_B\). \revise{Additional results using MobileNet-v2 are also included in Appendix~\ref{sec:mobilenet}}. Each client randomly draws 10\% to 90\% of its data from \(\mathcal{D}_A\)and the remainder from \(\mathcal{D}_B\) with unbalanced class \citep{marfoq2021federated}, image rotation \citep{ruan2022fedsoft}, or both. We thus create a portion of clients with significantly unbalanced data and guarantees the unique distribution of each client. We follow \citet{ruan2022fedsoft} and \citet{marfoq2021federated} for other parameter settings. Details are described in the Appendix \ref{sec:exsim}. The test accuracy is evaluated on each client's local test dataset, which is unseen during training.

\textbf{Client communications.} Unless specified, the client graph is a connected Erdős–Rényi (ER) random graph \citep{erdos1960evolution} with an average degree from 5 to 12; more details are in Appendix~\ref{sec:exsim}. %\carlee{where? I couldn't find them}. 
To avoid the label switching problem \citep{stephens2000dealing}, we calculate the cosine similarities of the model parameters received from other clients to ensure the consensus of the cluster.

\textbf{Baselines.} We compare \textbf{\algname}~with: (i) centralized and decentralized \textbf{FedAvg} \citep{mcmahan2017communication}; (ii) centralized and decentralized \textbf{FedEM} \citep{marfoq2021federated}, a prior soft clustering method; (iii) centralized and  decentralized versions of \textbf{FedSoft} \citep{ruan2022fedsoft}, which also uses soft clustering; (iv) centralized and decentralized \textbf{IFCA} \citep{ghosh2020efficient} using hard clustering; (v) centralized and decentralized \textbf{pFedMe} \citep{t2020personalized}, another state-of-the-art FL personalization approach without clustering; and (vi) \textbf{local} training on local dataset only.

\textbf{Additional results} will be included in the appendix due to page limit. These include:

\begin{itemize}
    \item We conduct ablation studies to analyze the impact of different factors on the performance of \textbf{FedSPD}, specifically evaluating:
    \begin{itemize}
        \item The influence of the number of local training epochs in Section~\ref{sec:fl_localep};
        \item The contribution of the final training phase in Section~\ref{sec:fl_final};
        \item The impact of the number of clusters in Section~\ref{sec:fl_cluster}; and
        \item More details of the effect of network connectivity in Section~\ref{sec:fl_edcon}. We also show how the dynamic network topology influences the performance of \textbf{\algname}.
    \end{itemize}
    \item In Section \ref{sec:IDA}, we evaluate the performance of \textbf{FedSPD} under a more challenging setting where the total amount of data is imbalanced across clients in addition to the data heterogeneity across clusters.
    \item To explore the potential for enhancing privacy guarantees in DFL, we incorporate Differential Privacy into \textbf{FedSPD} and present the results in Section~\ref{sec:dp}.
    \item For most experiments on the CIFAR-10/CIFAR-100 datasets, we adopt the same CNN model used by~\cite{ruan2022fedsoft} to ensure a fair comparison. To further assess the scalability of \textbf{FedSPD}, we evaluate its performance using a more complex architecture, MobileNet-V2, in Section~\ref{sec:mobilenet}.
\end{itemize}

\subsection{Comparison with Baselines}
We first compare our method with other decentralized personalized methods. Our results on EMNIST, CIFAR-10, and CIFAR-100 are shown in Table \ref{tab:test-cfl} and Table \ref{tab:test-dfl}. \textbf{\algname}~outperforms other DFL methods in most cases, approaching the accuracy of CFL. The centralized methods still outperform decentralized methods, as expected from prior literature \citep{sun2023mode}. However, decentralized methods offer advantages such as lower communication traffic and increased robustness, as they do not rely on a single point of failure like a centralized server. 

\begin{figure}[t]
    \centering
    \includegraphics[width=0.90\linewidth]{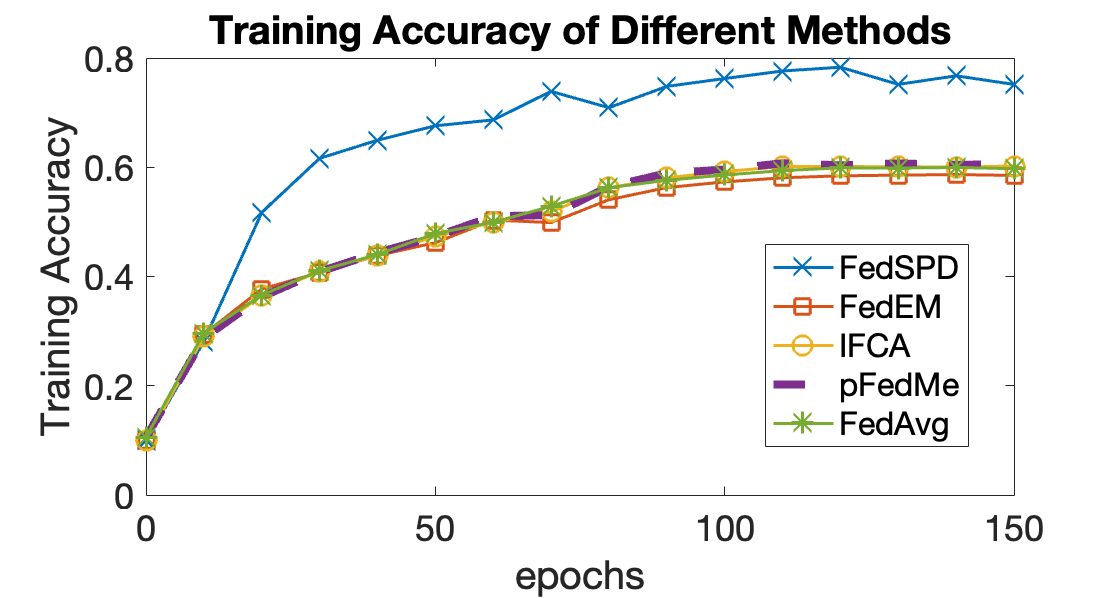}
\caption{\sl Training accuracy of different DFL methods versus number of epochs on CIFAR-10 ($N=25$). \textbf{\algname}~ converges faster in terms of training accuracy compared to all other DFL methods.
}
\label{fig:CMTA}
\end{figure}

Figure \ref{fig:CMTA} shows the training accuracy versus number of epochs on the CIFAR-10 dataset. \textbf{\algname}~\textit{converges faster} than all other DFL algorithms in terms of training accuracy. \new{This shows that each of the clusters in \textbf{\algname} does converge as desired. Note that compared to \textbf{FedEM}, another soft clustering method, our \textbf{\algname} needs half the communication cost, since \textbf{FedEM} clients exchange the information of all $S = 2$ clusters.}

\begin{table*}[htbp]
          \centering
\begin{tabular}
{|c|c|c|c|c|c|c|}
\hline
\multicolumn{1}{|c|}{} & \multicolumn{1}{|c|}{DFL} & \multicolumn{5}{|c|}{CFL} \\ \hline
%\hline  
Dataset & \textbf{FedSPD} & FedEM & IFCA & FedAvg & FedSoft & pFedMe \\
\hline EMNIST & 83.07 & 88.83 & 89.42 & 88.81 & 84.97 & 90.95 \\
\hline CIFAR-10 & 68.72 & 79.64 & 79.52 & 79.36 & 76.62 & 79.43 \\
\hline CIFAR-100 & 40.38 & 44.25 & 43.91 & 43.11 & 39.76 & 8.74\footnotemark \\
\hline
\end{tabular}
\caption{\sl \textbf{\algname}~has comparable test accuracy to CFL algorithms. Accuracy in percentage (\%) }
    \label{tab:test-cfl}
\end{table*}

\begin{table*}[htbp]
          \centering
\begin{tabular}
{|c|c|c|c|c|c|c|c|}
\hline
\multicolumn{1}{|c|}{} & \multicolumn{6}{|c|}{DFL} & \multicolumn{1}{|c|}{}\\ \hline
%\hline  
Dataset & \textbf{FedSPD} & FedEM & IFCA & FedAvg & FedSoft & pFedMe & Local\\
\hline EMNIST & 83.07 & 80.47 & \textbf{83.88} & 78.61 & 74.30 & 81.16 & 56.91\\
\hline CIFAR-10 & \textbf{68.72} & 50.45 & 52.39 & 49.21 & 42.38 & 49.48 & 41.82\\
\hline CIFAR-100 & \textbf{40.38} & 18.59 & 17.18 & 17.20 & 13.17 & 18.27 & 13.31\\
\hline
\end{tabular}
\caption{\sl \textbf{\algname}~achieves higher test accuracy than other DFL algorithms in most cases. Accuracy in percentage (\%) }
    \label{tab:test-dfl}
\end{table*}

\begin{figure}[t]
    \centering
    \includegraphics[width=0.90\linewidth]{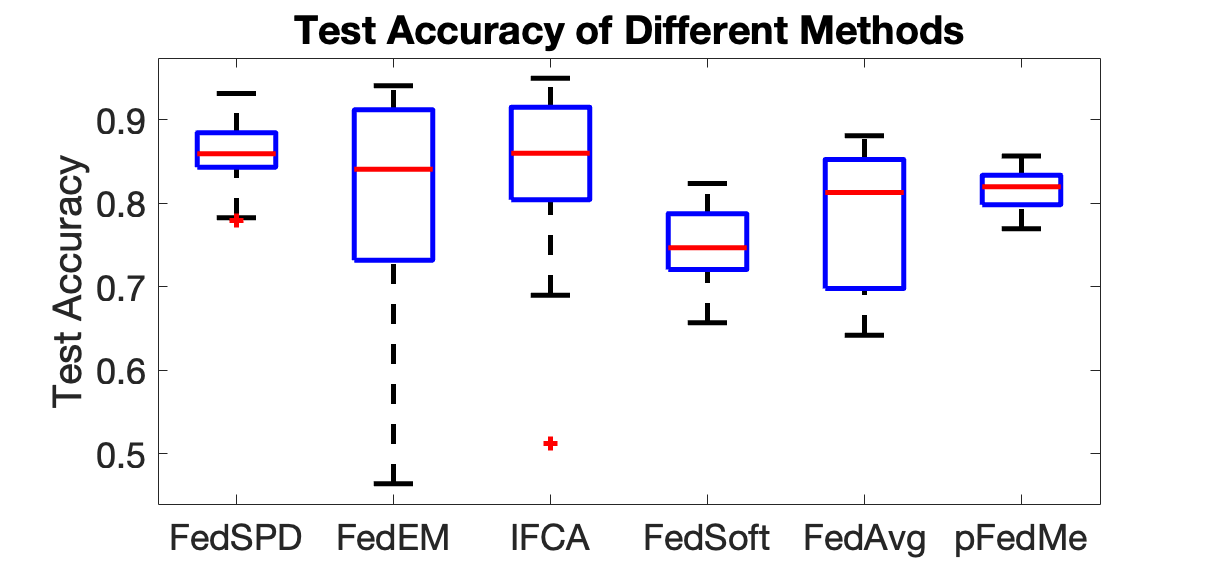}
\caption{\sl Box-plot for accuracy across clients on EMNIST dataset. \textbf{\algname}~has much lower variance in test accuracy across clients.}
\label{fig:BP}
\end{figure}

To guarantee the \textbf{fairness across clients}, we show the box plot of the final test accuracy across different clients on EMNIST in Figure \ref{fig:BP}. \textbf{\algname}~has much \textit{less variance in accuracy} across different clients compared to all baselines except \textbf{pFedMe}, validating that its improvement in average accuracy does not come from high accuracy in a few clients. 

\subsection{Effects of Network Connectivity}
\label{sec:sim_nc}
\begin{figure}[t]
    \centering
    \includegraphics[width=0.90\linewidth]{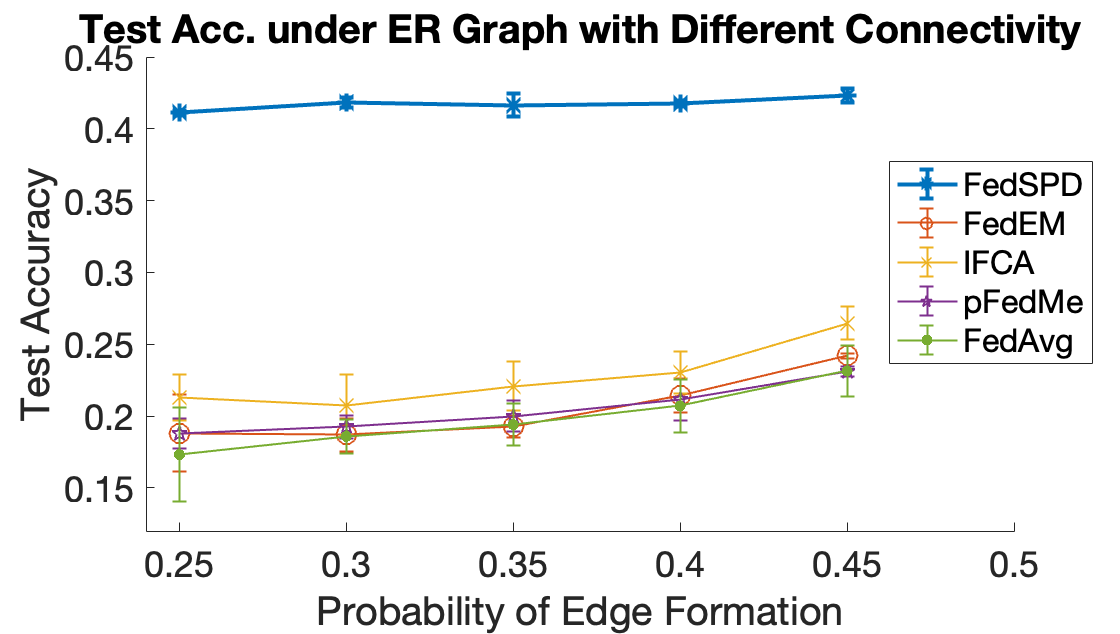}
\caption{\sl Test accuracy of different methods under different connectivity levels of an ER Random Graph on the CIFAR-100 dataset ($N=15$).  \textbf{\algname}~shows consistently high test accuracies compare to other DFL methods.}
\label{fig:CFCN}
\end{figure}

In this section, we investigate how the performance varies with the connectivity of the client network. Figure \ref{fig:CFCN} shows the test accuracy of different DFL methods under varying client connectivity on the CIFAR-100 dataset using the ER Random Graph, averaged over three experimental runs. \textbf{\algname}~consistently shows the highest test accuracies, though other methods' performance begins to increase as the graph becomes more connected (a higher probability of link formation). 

Tables \ref{tab:p2-MN} and \ref{tab:p2-EM} show the test accuracy of \textbf{\algname}~in different types of networks and connectivity levels. We use three different network topologies: the ER Random Graph; the Barabasi-Albert (BA) Model \citep{albert2002statistical} with preferential attachment representing the network following the power law; and the Random Geometrical Graph (RGG) \citep{penrose2003random}, which is often used in wireless communication and IoT scenarios that have high clustering effects.  We observe that the final test accuracy does not vary significantly across different network topologies and levels of connectivity in MNIST. In EMNIST, the test accuracy slightly increases when the average degree increases. The test accuracy is more stable in RGG under different connectivity, which we conjecture is due to RGG's highly clustered nature. Thus, as long as the network is connected, \textbf{\algname}~performs well in both high and low connectivity scenarios and across various types of networks. As we expect from Theorem~\ref{thm:4}, \textbf{\algname}~converges regardless of the network topology.

\subsection{Communication Overhead}

\textbf{\algname}~requires transmitting 50\% less data compared to \textbf{FedEM} (in the case $S=2$) since only a single model is transmitted by each client. As the number of clusters $S$ increases, our communication volume advantage grows. Compared to the decentralized versions of \textbf{FedAvg} and \textbf{FedSoft}, \textbf{\algname}~requires each client to send the same volume of data (equivalent to one model's parameters) in each round. \revise{However, since \textbf{\algname}~only requires clients to send their local models to their neighbors training models for the same cluster, the number of clients to which each client communicates in \textbf{\algname}~is smaller than in algorithms like \textbf{FedAvg} and \textbf{FedSoft}, which requires each client to send its local model to \textit{all} of its neighbors. \textbf{\algname}, \textbf{FedAvg} and \textbf{FedSoft} thus have comparable communication overhead if clients use multicast communication, but if they use point-to-point communication, \textbf{\algname}~will require less communication than \textbf{FedAvg} and \textbf{FedSoft} with full participation, due to having fewer recipients per client.}

\footnotetext{The centralized pFedMe on CIFAR-100 does not converge in the various settings of hyperparameters we tried.}

\subsection{Discussion}

As shown in Table \ref{tab:test-cfl} and Table \ref{tab:test-dfl}, local learning performs the worst among all algorithms, validating that all other methods benefit from exchanging information between clients to learn a better model. Among the DFL algorithms, \textbf{FedAvg}, the only one without personalization, typically performs the worst, indicating that personalization is beneficial in non-iid data distributions, as we would intuitively expect. However, an exception is observed with the \textbf{FedSoft} algorithm. In the CIFAR-10 and CIFAR-100 datasets, \textbf{FedSoft} performs poorly, nearing the accuracy of local training. We conjecture that this is due to the way \textbf{FedSoft} aggregates models, making it difficult to learn the correct cluster centers in a low-connectivity network, leading to suboptimal performance. Our \textbf{\algname} designs a new model update method to avoid such an issue. More detailed discussion comparing \textbf{\algname} and \textbf{FedSoft} can be found in Section \ref{sec:comp}.

\begin{table}[h!]
  \centering
\begin{tabular}
{|p{1.8cm}|p{0.75cm}|p{0.75cm}|p{0.75cm}|p{0.75cm}|p{0.75cm}|}
\hline  Avg. Degree & 6 & 8 & 10 & 12 & 14 \\
\hline ER & 92.86 & 92.93 & 93.37& 93.31& 93.26 \\
\hline BA & 93.06 & 92.58 & 92.56& 92.87& 93.17 \\
\hline RGG & 92.86 & 92.61 & 92.84& 93.49& 92.97 \\
\hline
\end{tabular}
\caption{\sl \textbf{\algname}~shows consistently high test accuracies on MNIST data for $N=50$ clients across different client network topologies.}
  \label{tab:p2-MN}
\end{table}

\begin{table}[h!]
  \centering
\begin{tabular}
{|p{1.8cm}|p{1.05cm}|p{1.05cm}|p{1.05cm}|p{1.05cm}|}
\hline  Avg. Degree & 8 & 12 & 16 & 20  \\
\hline ER & 79.79 & 82.26 & 84.28 & 84.49 \\
\hline BA & 79.45 & 82.13 & 84.58 & 84.73 \\
\hline RGG & 82.26 & 83.49 & 84.06 & 84.08\\
\hline
\end{tabular}
\caption{\sl \textbf{\algname}~shows consistently high test accuracies on EMNIST data for $N=50$ clients across different client network topologies.}
  \label{tab:p2-EM}
\end{table}

%% file: sections/conclusion.tex
\section{Conclusion}\label{sec:conclusion}
We propose \textbf{\algname}, a soft clustering approach that enables federated training of personalized models in a decentralized setting. \textbf{\algname}~models each FL client's data as a mixture of cluster distributions and aims to learn a distinct model for each cluster. In the final phase, all models are aggregated and further personalized for each client. Importantly, \textbf{\algname}~requires each client to train only one cluster model per training round, ensuring scalability with the number of clusters, and works well when communication resource is limited. We theoretically demonstrate that \textbf{\algname}~can achieve consensus within each cluster. Our experiments on real-world datasets show that \textbf{\algname}~outperforms previous algorithms for personalized, decentralized FL and performs well even in \textit{low-connectivity} networks. For future extensions, this work can serve as a foundation for various applications, such as environmental monitoring in IoT, object identification in mixed reality, or autonomous driving, all of which benefit from the low latency of direct communication and collaborative learning across adjacent devices with similar data.

%% file: sections/appendix.tex
\section{Proof of the Theorems}\label{sec:proof}

\input{sections/thm1}
\input{sections/thm2}
\input{sections/thm3}
\input{sections/thm4}

\section{Simulation Details and Additional Simulations}\label{sec:exsim}
\input{sections/Extrasim_new}

\input{sections/compare}

%% file: sections/thm1.tex
\subsection{Proof of Theorem \ref{thm:1}}
Without loss of generality, we select a single cluster, cluster 1 for analysis; the same analysis applies to the other $S - 1$ clusters. For readability, we eliminate the subscription indicating the cluster number 1. Consider each client running single step of SGD, we use $n$ to indicate the number of clients selected to update this cluster and $n_1$ and $n_0$ to indicate the number of clients using the correct data and incorrect data, respectively
 \new{ (i.e. the data is drawn from this selected cluster is consider a correct data.)}, so that $n_1 + n_0 = n$. $S$ indicates the set of the client selected to update this cluster, $S*$ indicates the set of clients using the correct data, and $\overline{S*}$ indicates the set of clients using the incorrect data.
\begin{lemma}\label{lm:ds}
    (Doubly-stochastic weight matrix preserves the average) At the communication step, if the model of each client in the network is updated according to a doubly-stochastic weight matrix $\mathbf{W}^t$ then the average after the communication step remains the same. Formally, we have:
    \begin{equation}
        \mathbf{C}^{t+1}\frac{\mathbf{1}\mathbf{1}^T}{N} = \mathbf{C}^{t}\mathbf{W}^{t}\frac{\mathbf{1}\mathbf{1}^T}{N} = \mathbf{C}^{t}\frac{\mathbf{1}\mathbf{1}^T}{N}
    \end{equation}
\end{lemma}

From Lemma \ref{lm:ds}, we can write the left-hand side of Theorem \ref{thm:1} as:

\begin{equation}
\begin{aligned}
\left\|\overline{\mathbf{c}}^{(t+1)}-\mathbf{c}^{\star}\right\|^2= & \left\|\overline{\mathbf{c}}^{(t)}-\frac{\eta_t}{n} \sum_{i=1}^n \nabla F_i\left(\mathbf{c}_i^{(t)}, D_i^{(t)}\right)-\mathbf{c}^{\star}\right\|^2 \\
= & \left\|\overline{\mathbf{c}}^{(t)}-\mathbf{c}^{\star}-\frac{\eta_t}{n} \sum_{i \in S \cap S*} \nabla F_i\left(\mathbf{c}_i^{(t)}\right) -\frac{\eta_t}{n} \sum_{i \in S \cap \overline{S*}} \nabla F_i\left(\mathbf{c}_i^{(t)}\right)\right\|^2 \\
= & \left\|\overline{\mathbf{c}}^{(t)}-\mathbf{c}^{\star}-\frac{\eta_t}{n} \sum_{i \in S \cap S*} \nabla F_i\left(\mathbf{c}_i^{(t)}\right)\right\|^2+\left\|\frac{\eta_t}{n} \sum_{i \in S \cap \overline{S*}} \nabla F_i\left(\mathbf{c}_i^{(t)}\right)\right\|^2\\
& -\frac{2 \eta_t}{n}\left\langle\overline{\mathbf{c}}^{(t)}-\mathbf{c}^{\star}-\frac{\eta_t}{n} \sum_{i \in S \cap S*} \nabla F_i\left(\mathbf{c}_i^{(t)}\right), \sum_{i \in S \cap \overline{S*}} \nabla F_i\left(\mathbf{c}_i^{(t)}\right)\right\rangle
\end{aligned}
\end{equation}

We let the first and the second term on the right-hand side as $\| T_1 \|^2$ and $\| T_2 \|^2$ respectively. Thus the above equation can be written as:

\begin{equation}
\left\|\overline{\mathbf{c}}^{(t+1)}-\mathbf{c}^{\star}\right\|^2= \| T_1 \|^2 + \| T_2 \|^2 +2 \left\langle T_1, T_2 \right\rangle \leq (1+\alpha)\| T_1 \|^2 + (1+\alpha^{-1})\| T_2 \|^2
\end{equation}

for all $\alpha>0$.

The $T_1$ part is the typical decentralized SGD items. Inspired by \citep{koloskova2020unified}, we write $T_1$ as: 

\begin{equation}
\begin{aligned}
    \left\|\overline{\mathbf{c}}^{(t)}-\mathbf{c}^{\star}-\frac{\eta_t}{n} \sum_{i \in S \cap S*} \nabla F_i\left(\mathbf{c}_i^{(t)}\right)\right\|^2 & \leq \left\|\overline{\mathbf{c}}^{(t)}-\mathbf{c}^{\star}\right\|^2+\eta_t^2\frac{n_1^2}{n^2} \underbrace{\left\|\frac{1}{n_1} \sum_{i=1}^{n_1} \nabla f_i\left(\mathbf{c}_i^{(t)}\right)\right\|^2}_{T_{11}} \\
    & + 2 \eta_t\frac{n_1}{n}\left\langle\underbrace{\overline{\mathbf{c}}^{(t)}-\mathbf{c}^{\star}, \frac{-1}{n_1} \sum_{i=1}^{n_1} \nabla f_i\left(\mathbf{c}_i^{(t)}\right)}_{T_{12}}\right\rangle + \eta_t^2 v^2
\end{aligned}
\end{equation}

We can bound $T_{11}$ and $T_{12}$ separately as:

\begin{equation}
\begin{aligned}
T_{11} & =\left\|\frac{1}{n_1} \sum_{i=1}^{n_1}\left(\nabla f_i\left(\mathbf{c}_i^{(t)}\right)-\nabla f_i\left(\overline{\mathbf{c}}^{(t)}\right)+\nabla f_i\left(\overline{\mathbf{c}}^{(t)}\right)-\nabla f_i\left(\mathbf{c}^{\star}\right)\right)\right\|^2 \\
& \leq \frac{2}{n_1} \sum_{i=1}^{n_1}\left\|\nabla f_i\left(\mathbf{c}_i^{(t)}\right)-\nabla f_i\left(\overline{\mathbf{c}}^{(t)}\right)\right\|^2+2\left\|\frac{1}{n} \sum_{i=1}^{n_1} \nabla f_i\left(\overline{\mathbf{c}}^{(t)}\right)-\frac{1}{n} \sum_{i=1}^{n_1} \nabla f_i\left(\mathbf{c}^{\star}\right)\right\|^2 \\
& =\frac{2 L^2}{n_1} \sum_{i=1}^{n_1}\left\|\mathbf{c}_i^{(t)}-\overline{\mathbf{c}}^{(t)}\right\|^2+4 L\left(f\left(\overline{\mathbf{c}}^{(t)}\right)-f(\mathbf{c}^{\star})\right)
\end{aligned}
\end{equation}

\begin{equation}
\begin{aligned}
-T_{12} & =-\frac{1}{n_1} \sum_{i=1}^{n_1}\left[\left\langle\overline{\mathbf{c}}^{(t)}-\mathbf{c}_i^{(t)}, \nabla f_i\left(\mathbf{c}_i^{(t)}\right)\right\rangle+\left\langle\mathbf{c}_i^{(t)}-\mathbf{c}^{\star}, \nabla f_i\left(\mathbf{c}_i^{(t)}\right)\right\rangle\right] \\
& \leq-\frac{1}{n_1} \sum_{i=1}^{n_1}\left[f_i\left(\overline{\mathbf{c}}^{(t)}\right)-f_i\left(\mathbf{c}_i^{(t)}\right)-\frac{L}{2}\left\|\overline{\mathbf{c}}^{(t)}-\mathbf{c}_i^{(t)}\right\|^2+f_i\left(\mathbf{c}_i^{(t)}\right)-f_i\left(\mathbf{c}^{\star}\right)+\frac{\mu}{2}\left\|\mathbf{c}_i^{(t)}-\mathbf{c}^{\star}\right\|^2\right] \\
& \leq -\left(f\left(\overline{\mathbf{c}}^{(t)}\right)-f\left(\mathbf{c}^{\star}\right)\right)+\frac{L+\mu}{2n_1} \sum_{i=1}^{n_1}\left\|\overline{\mathbf{c}}^{(t)}-\mathbf{c}_i^{(t)}\right\|^2-\frac{\mu}{4}\left\|\overline{\mathbf{c}}^{(t)}-\mathbf{c}^{\star}\right\|^2
\end{aligned}
\end{equation}

Now we deal with $T_2$. From \citep{ruan2022fedsoft} and \citep{ghosh2020efficient} we have the following Lemma:

\begin{lemma}\label{lm:ep}
    (Mis-classified probability) For a data point belongs to cluster $j$, the probability of error classification $\mathbb{P}(\epsilon^{j, j'})$ to cluster $j' \neq j$ can be bound as:
    \begin{equation}
        \mathbb{P}(\epsilon^{j, j'}) \leq \frac{c_1}{\alpha_0^2\delta^4}
    \end{equation}
    And by union bound, the error probability is bounded as:
    \begin{equation}
        \mathbb{P}(\overline{\epsilon}) \leq \frac{c_1 S}{\alpha_0^2\delta^4}
    \end{equation}
    The expected number of clients using wrong cluster of data is bounded as:
    \begin{equation}
        \mathbb{E}[S \cap \overline{S*}] \leq \frac{c_1 N}{\alpha_0^2\delta^4} = \epsilon_N
    \end{equation}
    for some constant $c_1$.
    We define this bound as $\epsilon_N$
    
\end{lemma}

Inspired by \citep{ghosh2020efficient}, define $T_{2k}$ as the clients selecting the mis-classified data points that should be belongs to cluster $k$ where $k \neq 1$(The correct cluster). That is:

\begin{equation}
\begin{aligned}
T_{2k} = \sum_{i\in S \cap \overline{S*} \cap S_k*}\nabla F_i(\mathbf{c}_i)
\end{aligned}
\end{equation}

For each $T_{2k}$, we use $n_k$ to indicate the number of clients using mis-classified data that should be belongs to cluster $k$. We have:

\begin{equation}
\begin{aligned}
T_{2k} = \sum_{i=1}^{n_k}\nabla F_i^k(\mathbf{c}_i) + \sum_{i=1}^{n_k}\nabla F_i (\mathbf{c}_i) - \nabla F_i^k(\mathbf{c}_i)
\end{aligned}
\end{equation}

Taking the expectation and by Markov's inequality:

\begin{equation}
\begin{aligned}
\|T_{2k}\| & = \left\| \sum_{i=1}^{n_k}\nabla F_i^k(\mathbf{c}_i) + \sum_{i=1}^{n_k}\nabla F_i (\mathbf{c}_i) - \nabla F_i^k(\mathbf{c}_i) \right\| \\
& \leq 3 n_k L + \frac{\sqrt{n_k} v}{\theta_1} 
\end{aligned}
\end{equation}

For any $\theta_1 \in (0, 1)$ with probability equal or greater than $1-\theta_1$. The above used Lemma \ref{lm:ep}, Assumption \ref{as:5} and Assumption \ref{as:6} and the Markov inequality.

Using the union bound we see that $T_2 = \sum_k T_{2k}$ is bounded as the following with probability greater or equal to $1-(S-1)\theta_1-\theta_2$:

\begin{equation}
\begin{aligned}
\|T_{2}\|^2 & = \| \sum_{k=2}^S T_{2k}\|^2 \leq (S-1) \sum_{k=2}^S  \|T_{2k}\|^2 \\
& \leq \frac{18L^2\epsilon_N^2}{\theta_2^2} + \frac{2\epsilon_N (S-1) v^2}{\theta_1^2 \theta_2}
\end{aligned}
\end{equation}

When $\sum_{k=2}^S n_k \leq \frac{\epsilon_N}{\theta_2}$ with probability at least $1-\theta_2$. %: \carlee{this sentence does not make sense}

Combining the above three terms and Lemma \ref{lm:ep}, we have:

\begin{equation}
\begin{aligned}
\mathbb{E}\left\|\overline{\mathbf{c}}^{(t+1)}-\mathbf{c}^{\star}\right\|^2 & \leq (1- \eta_t \mu + \eta_t \mu\frac{\epsilon_N}{n})\left\|\overline{\mathbf{c}}^{(t)}-\mathbf{c}^{\star}\right\|^2 + \frac{18L^2\epsilon_N^2 \eta_t^2}{n^2} + \frac{2\epsilon_N (S-1) v^2 \eta_t^2}{n^2} + \eta_t^2v^2 \\
& + \frac{\eta_t (L+\mu)}{n}\sum_{i=1}^{n_1}\left\|\overline{\mathbf{c}}^{(t)}-\mathbf{c}_i^{(t)}\right\|^2 + \left(\frac{4\eta_t^2(n-\epsilon_N)^2 L}{n^2} + 2\eta_t - \frac{2\eta_t \epsilon_N}{n}\right) \left(f\left(\overline{\mathbf{c}}^{(t)}\right)-f\left(\mathbf{c}^{\star}\right)\right)
\end{aligned}
\end{equation}

%% file: sections/thm2.tex
\subsection{Proof of Theorem \ref{thm:2}}
For time $t-1$, after the local updating round, the cluster parameters can be expressed as:
\begin{equation}
    \mathbf{C}^{t-1'} = \mathbf{C}^{t-1} - \eta_t \mathbf{G}^{t-1}
\end{equation}

After the communication round, the parameters can be expressed as:
\begin{equation}
    \mathbf{C}^{t} = \mathbf{C}^{t-1'}\mathbf{W}^{t-1} = \mathbf{C}^{t-1}\mathbf{W}^{t-1} - \eta_t \mathbf{G}^{t-1}\mathbf{W}^{t-1}
\end{equation}

Thus, recursively expanding the parameters at time $t$ back to $l\beta$, we can get the final form:
\begin{equation}
\begin{aligned}
\mathbf{C}^{t} & = \mathbf{C}^{l\beta}\prod_{m=l\beta}^{t-1}\mathbf{W}^{m}-\sum_{m=l\beta}^{t-1}\left(\eta_t\mathbf{G}^{m} \prod_{r=t-1}^{m}\mathbf{W}^{r}\right)
\end{aligned}
\end{equation}

%% file: sections/thm3.tex
\subsection{Proof of Theorem \ref{thm:3}}
Following the same flow of Lemma 9 in \citep{koloskova2020unified}, applying Theorem \ref{thm:3}, we have for all $\alpha > 0$:

\begin{equation}
\begin{aligned}
\mathbb{E} \| \mathbf{C}^{t}-\overline{\mathbf{C}}^{t} \|_F^2 & = N \mathbf{E}_t \leq \mathbb{E}\left\|\mathbf{C}^{(m \beta)} \prod_{i=t-1}^{m \beta} \mathbf{W}^{(i)}-\bar{\mathbf{C}}^{\left(m \beta \right)}+\sum_{j=m \beta}^{t-1} \eta_j \nabla \mathbf{F}\left(\mathbf{C}^{(j)}\right) \prod_{i=t-1}^j \mathbf{W}^{(i)}\right\|_F^2 \\
& \leq \mathbb{E}\left\|\mathbf{C}^{(m \beta)} \prod_{i=t-1}^{m \beta} \mathbf{W}^{(i)}-\bar{\mathbf{C}}^{\left(m \beta \right)}+\sum_{j=m \beta}^{t-1} \eta_j \left(\nabla \mathbf{F}\left(\mathbf{C}^{(j)}\right) - \nabla \mathbf{F}\left(\mathbf{C}^{\star}\right) + \nabla \mathbf{f}\left(\mathbf{C}^{\star}\right) \right) \prod_{i=t-1}^j \mathbf{W}^{(i)}\right\|_F^2 \\
& + \left \| \sum_{j=m \beta}^{t-1} \eta_j \left( \nabla \mathbf{F}\left(\mathbf{C}^{\star}\right) + \nabla \mathbf{f}\left(\mathbf{C}^{\star}\right) \right) \prod_{i=t-1}^j \mathbf{W}^{(i)}\right\|_F^2 \\
& \leq (1+\alpha) \mathbb{E}\left\|\mathbf{C}^{(m \beta)} \prod_{i=t-1}^{m \beta} \mathbf{W}^{(i)}-\bar{\mathbf{C}}^{\left(m \beta \right)}\right \|_F^2 \\
& + (1+\alpha^{-1})\mathbb{E}\left\|\sum_{j=m \beta}^{t-1} \eta_j \left(\nabla \mathbf{F}\left(\mathbf{C}^{(j)}\right) - \nabla \mathbf{F}\left(\mathbf{C}^{\star}\right) + \nabla \mathbf{f}\left(\mathbf{C}^{\star}\right) \right) \prod_{i=t-1}^j \mathbf{W}^{(i)}\right\|_F^2 \\
& + \left \| \sum_{j=m \beta}^{t-1} \eta_j \left( \nabla \mathbf{F}\left(\mathbf{C}^{\star}\right) + \nabla \mathbf{f}\left(\mathbf{C}^{\star}\right) \right) \prod_{i=t-1}^j \mathbf{W}^{(i)}\right\|_F^2 \\
\end{aligned}    
\end{equation}

Using Assumption \ref{as:7}, the above can be further simplified:

\begin{equation}
\begin{aligned}
\mathbb{E} \| \mathbf{C}^{t}-\overline{\mathbf{C}}^{t} \|_F^2 & \leq (1+\alpha) (1-p) \mathbb{E}\left\|\mathbf{C}^{(m \beta)} -\overline{\mathbf{C}}^{(m \beta)} \right \|_F^2 \\
& + (1+\alpha^{-1}) 2 \beta \sum_{j=m \beta}^{t-1} \eta_j^2 \mathbb{E}\left \|\left(\nabla \mathbf{F}\left(\mathbf{C}^{(j)}\right) - \nabla \mathbf{F}\left(\mathbf{C}^{\star}\right) + \nabla \mathbf{f}\left(\mathbf{C}^{\star}\right) \right) \right \|_F^2 \\
& + \sum_{j=m \beta}^{t-1} \eta_j^2 \mathbb{E}\left \|\left( \nabla \mathbf{F}\left(\mathbf{C}^{\star}\right) + \nabla \mathbf{f}\left(\mathbf{C}^{\star}\right) \right) \right \|_F^2 \\
& \leq (1+\alpha) (1-p) \mathbb{E}\left\|\mathbf{C}^{(m \beta)} -\overline{\mathbf{C}}^{(m \beta)} \right \|_F^2 \\
& + (1+\alpha^{-1}) 2 \beta \sum_{j=m \beta}^{t-1} \eta_j^2 \mathbb{E}\left \|\left(\nabla \mathbf{F}\left(\mathbf{C}^{(j)}\right) - \nabla \mathbf{F}\left(\mathbf{C}^{\star}\right) + \nabla \mathbf{f}\left(\mathbf{C}^{\star}\right) \right) \right \|_F^2 \\
& + \sum_{j=m \beta}^{t-1} \eta_j^2 nv^2 \\
\end{aligned}    
\end{equation}

The expectation of the second term on the right-hand side can be bounded as:
\begin{equation}
\begin{aligned}
&\mathbb{E}\left\|\left(\nabla \mathbf{F}\left(\mathbf{C}^{(j)}\right)  - \nabla \mathbf{F}\left(\mathbf{C}^{\star}\right) + \nabla \mathbf{f}\left(\mathbf{C}^{\star}\right) \right) \right\|_F^2 \\
& = \mathbb{E}\left \|\left(\nabla \mathbf{F}\left(\mathbf{C}^{(j)}\right) - \nabla \mathbf{F}\left(\overline{\mathbf{C}}\right) + \nabla \mathbf{F}\left(\overline{\mathbf{C}}\right) - \nabla \mathbf{F}\left(\mathbf{C}^{\star}\right) + \nabla \mathbf{f}\left(\mathbf{C}^{\star}\right) \right) \right \|_F^2 \\
& \leq 3 \frac{n}{N} L^2 \|\mathbf{C}^{(j)} - \overline{\mathbf{C}}^{(j)}\|_F^2 + 3n \sigma^2 + 6nL(f(\overline{\mathbf{c}^{{j}}})-f(\mathbf{c}^{\star})) \\
& \leq 3 \frac{n}{N} L^2 \|\mathbf{C}^{(j)} - \overline{\mathbf{C}}^{(j)}\|_F^2 + 3n \sigma^2 + 6nL(f(\overline{\mathbf{c}^{{j}}})-f(\mathbf{c}^{\star}))
\end{aligned}    
\end{equation}

Putting the above equations together and setting a proper $\alpha$ to make the first term become $1-\frac{p}{2}$, similar to \citep{koloskova2020unified} with stepsize $\eta_j \leq \frac{p\sqrt{N}}{12\sqrt{2n}\beta L}$, we can get the desired bound:

\begin{equation}
\begin{aligned}
\mathbf{E}_t & \leq (1-\frac{p}{2}) \mathbf{E}_{m\beta} + \frac{p}{16\beta} \sum_{j=m\beta}^{t-1} \mathbf{E}_j + \frac{36Ln\beta}{pN} \sum_{j=m\beta}^{t-1} \eta_j^2 (f(\overline{\mathbf{c}^{{j}}})-f(\mathbf{c}^{\star})) \\
& + \left (  \frac{18\beta n}{Np} \sigma^2 + \frac{n}{N} v^2 \right ) \sum_{j=m\beta}^{t-1} \eta_j^2
\end{aligned}    
\end{equation}

%% file: sections/thm4.tex
\subsection{Proof of Theorem \ref{thm:4}}\label{sec:thm4-proof}
We adapted the following Lemma \ref{lm:cr} from \citep{koloskova2020unified}:

\begin{lemma}\label{lm:cr}
    (Simplify the Recursive Equations) For a bound of the cluster distance to the optimal $d_t = \mathbb{E}\| \overline{\mathbf{c}}^{(t)} - \mathbf{c}^{\star}\|^2$ in the following form:
    \begin{equation}
        d_{t+1} \leq\left(1-a \eta_t\right) d_t-b \eta_t e_t+c \eta_t^2+\eta_t B \mathbf{E}_t,
    \end{equation}
    
    and for any non-negative sequences $\{\mathbf{E}_t\}_{t \geq 0}, \{e_t\}_{t \geq 0}, \{\eta_t\}_{t \geq 0}$ that satisfy the following form:
    \begin{equation}
        \mathbf{E}_t \leq\left(1-\frac{p}{2}\right) \mathbf{E}_{m \beta}+\frac{p}{16 \beta} \sum_{j=m \beta}^{t-1} \mathbf{E}_j+D \sum_{j=m \beta}^{t-1} \eta_j^2 e_j+A \sum_{j=m \beta}^{t-1} \eta_j^2,
    \end{equation}

    then if the learning rate $\{ \eta_t^2 \}_{t \geq 0}$ and $\{ r_t \}_{t \geq 0}$ \new{are respectively} a $\frac{8\beta}{p}$-slow decreasing sequence and $\frac{16\beta}{p}$-slow increasing non-negative sequence, then for some constant $E > 0$ with learning rate $\eta_t \leq \frac{1}{16} \sqrt{\frac{p b}{D B \beta}}$ the following holds:
    \begin{equation}
    E \sum_{t=0}^T r_t \mathbf{E}_t \leq \frac{b}{2} \sum_{t=0}^T r_t e_t+64 B A \frac{\beta}{p} \sum_{t=0}^T r_t \eta_t^2
    \end{equation}

    By combining the above equations we have:

    \begin{equation}
        \frac{1}{2 R_T} \sum_{t=0}^T b r_t e_t \leq \frac{1}{R_T} \sum_{t=0}^T\left(\frac{\left(1-a \eta_t\right) r_t}{\eta_t} d_t-\frac{r_t}{\eta_t} d_{t+1}\right)+\frac{c}{R_T} \sum_{t=0}^T r_t \eta_t+\frac{64 B A}{R_T} \sum_{t=0}^T r_t \eta_t^2
    \end{equation}

    Where $R_T = \sum_{t=0}^T r_t$
\end{lemma}

Following the previous Lemma, we adapt Lemma 13 from \citep{koloskova2020unified} as the following Lemma \ref{lm:mr}

\begin{lemma}\label{lm:mr}
    (Main Recursion) The main recursion can be bounded as the following with a constant step-size $\eta_t = \eta < \frac{1}{h}$:
    \begin{equation}
    \frac{1}{2 R_T} \sum_{t=0}^T b e_t r_t+a d_{T+1} \leq \tilde{\mathcal{O}}\left(d_0 h \exp \left[-\frac{a(T+1)}{h}\right]+\frac{c}{a T}+\frac{B A}{a^2 T^2}\right)
    \end{equation}

    For the following two cases, tuning $\eta$ we have:
    If $\frac{1}{h} \geq \frac{\ln \left(\max \left\{2, a^2 d_0 T^2 / c\right\}\right)}{a T}$ $\eta$ is chosen to be equal to this value and that:

    \begin{equation}
        \tilde{\mathcal{O}}\left(a d_0 T \exp \left[-\ln \left(\max \left\{2, a^2 d_0 T^2 / c\right\}\right)\right]\right)+\tilde{\mathcal{O}}\left(\frac{c}{a T}\right)+\tilde{\mathcal{O}}\left(\frac{B A}{a^2 T^2}\right)=\tilde{\mathcal{O}}\left(\frac{c}{a T}\right)+\tilde{\mathcal{O}}\left(\frac{B A}{a^2 T^2}\right)
    \end{equation}

    If else choose $\eta = \frac{1}{h}$ and that:

    \begin{equation}
        \tilde{\mathcal{O}}\left(d_0 h \exp \left[-\frac{a(T+1)}{h}\right]+\frac{c}{h}+\frac{B A}{h^2}\right) \leq \tilde{\mathcal{O}}\left(d_0 h \exp \left[-\frac{a(T+1)}{h}\right]+\frac{c}{a T}+\frac{B A}{a^2 T^2}\right)
    \end{equation}

\end{lemma}

Using the above Lemma \ref{lm:cr}, Lemma \ref{lm:mr} and Theorem \ref{thm:1} and Theorem \ref{thm:3}, we can get the final bound.

%% file: sections/Extrasim_new.tex
\subsection{Experiment Details}
The following shows the detailed settings of our experiments. \new{We largely follow \cite{marfoq2021federated, ruan2022fedsoft} in our experiment settings.}

\subsubsection{MNIST/EMNIST Data}
Half of the dataset was selected to undergo a 90-degree rotation. Each client received the same amount of data, but the ratio of rotated to non-rotated data was set uniformly at random in the range from 10\% and 90\%. The number of clients was fixed at $N = 100$ for comparison with the baselines. A CNN (convolutional neural network) model was employed, consisting of two convolutional layers with kernel size and padding set to 5 and 2, respectively. Each convolutional layer was followed by a max-pooling layer with a kernel size of 2. After the convolutional layers, fully connected layers were used, with a dropout layer of size 50. The ReLU activation function was applied \new{to each convolution layer and fully-connected layer.} %\carlee{to each fully connected layer?}. 
All clients utilized SGD as the optimizer. The number of local epochs was set to 5, with the initial step having double the local epochs \new{to accelerate the initial learning, leading to a faster reduction in global loss.}
The initial learning rate was 5e-2, with a decay factor of 0.80. Training was carried out over 150 global epochs. Regarding network topology, unless otherwise specified, ER Random Graph with connecting probability $p=0.06$ and total number of clients $N=100$ were used in the experiments. The results in the table are averaged over five individual experiments.

\subsubsection{CIFAR-10 \& CIFAR-100 Data}
The dataset was divided into even and odd labels \new{by its number of label marked in the dataset}, and half of the data was randomly selected to undergo a 90-degree rotation. This process potentially created four different data distributions (rotated even, un-rotated even, rotated odd, un-rotated odd). Each client received an equal amount of data, but the proportion of odd-labeled and even-labeled data was randomly assigned, ranging uniformly at random from 10\% to 90\%. The number of clients was set to $N = 25$ for comparison with the baselines. A CNN model with four convolutional layers was used. The first two layers had a kernel size and padding of 5 and 2, respectively, while the last two layers had a kernel size and padding of 3 and 1, respectively. Each convolutional layer was followed by batch normalization. After the second and fourth convolutional layers, max-pooling with a kernel size of 2 and a dropout layer were applied. Following the convolutional layers, two fully connected layers with dropout and batch normalization were used, containing 1024 and 512 hidden neurons, respectively. The activation function was ReLU \new{on each layer.} All clients used SGD as the optimizer. The number of local epochs was set to 5, with the initial step doubling the local epochs. The initial learning rate was set to 5e-2, with a decay factor of 0.85. Training was conducted for 150 global epochs. Regarding network topology, unless otherwise specified, ER Random Graph with connecting probability $p=0.20$ and total number of clients $N=25$ were used in the experiments. The results in the table are averaged over two independent runs, except for FedSoft, which was only run once due to its extensive runtime. The results in the Figure are averaged over three independent runs with slightly lower number of clients $N=15$.

For the code for FedSPD, please refer to the \href{https://github.com/Anonymous-Submission-for-AISTATS/FedSPD_Anonymous_Submission}{Link}.

\subsection{Additional Simulation Results}
\subsubsection{Varying the Number of Local Epochs} \label{sec:fl_localep}
\begin{figure*}[h!]
\begin{subfigure}{0.32\linewidth}
    \centering
    \includegraphics[width=1.0\linewidth]{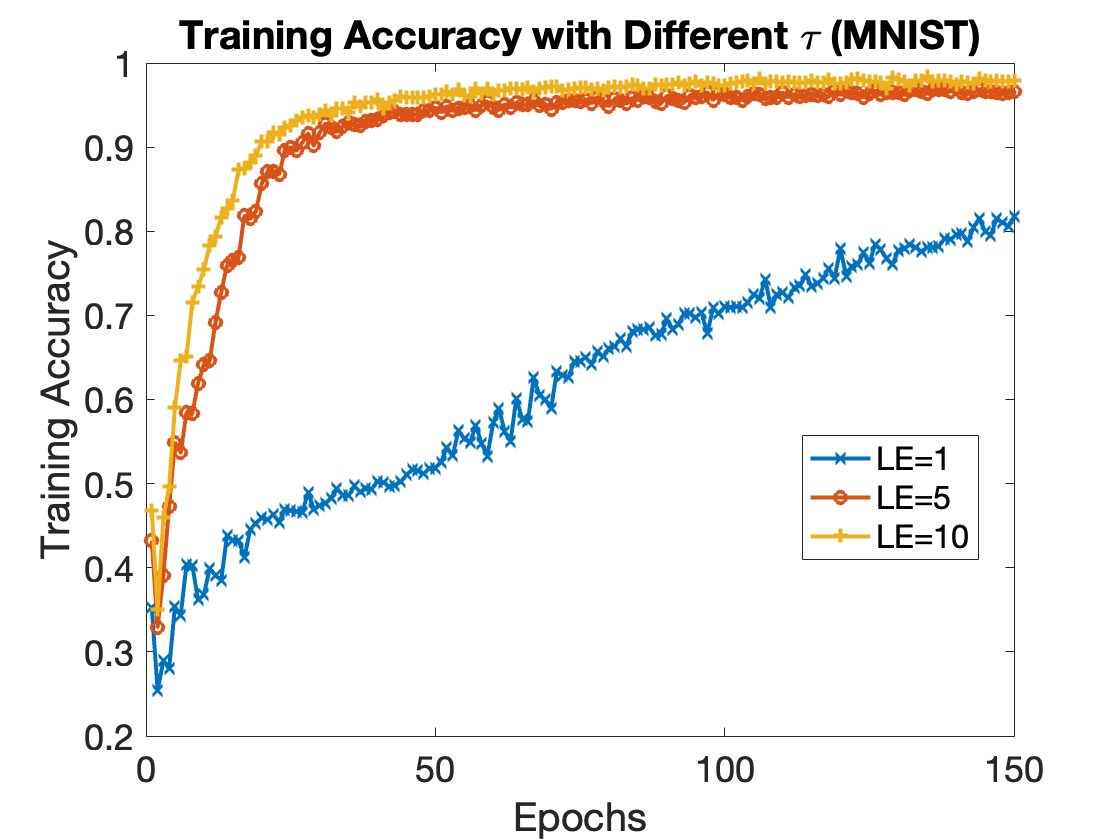}
    \caption{\sl Training accuracy on MNIST.}
    \label{fig:TA_MN}
\end{subfigure}
\hfill
\begin{subfigure}{0.32\textwidth}
    \centering
    \includegraphics[width=1.0\linewidth]{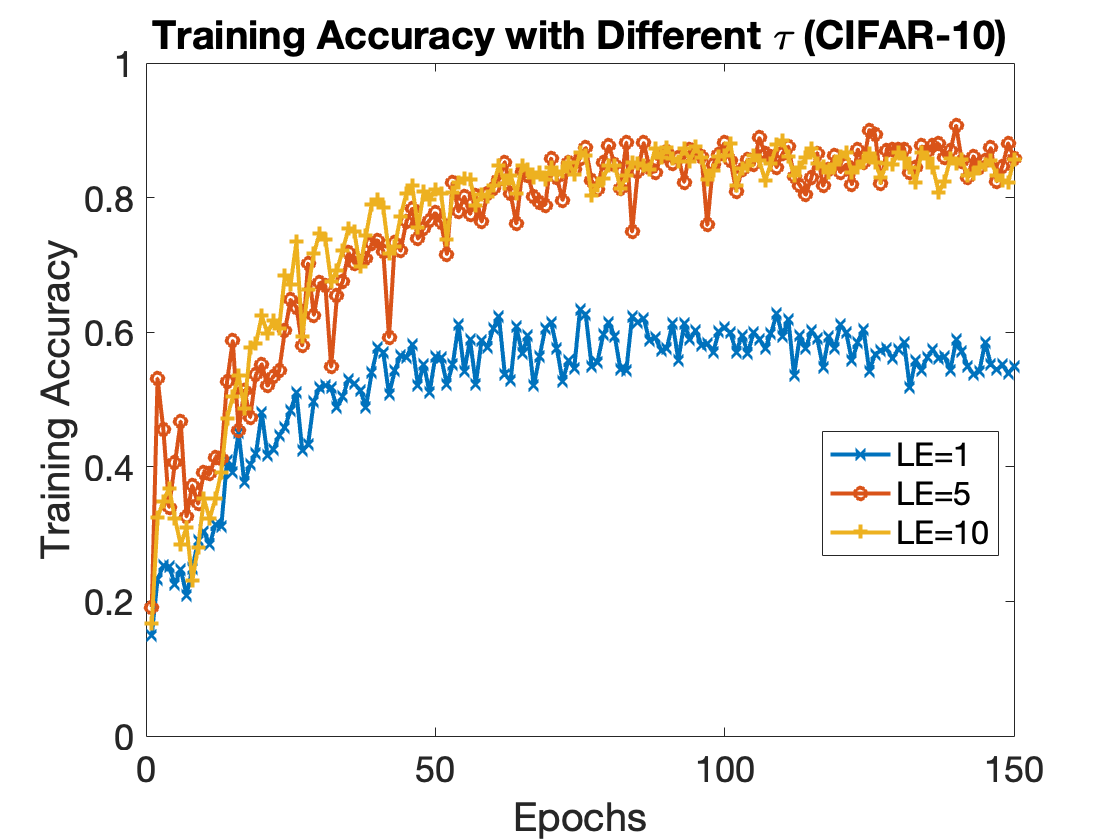}
    \caption{\sl Training accuracy on CIFAR-10.}
    \label{fig:TA_C10}
\end{subfigure}
\hfill
\begin{subfigure}{0.32\textwidth}
    \centering
    \includegraphics[width=1.0\linewidth]{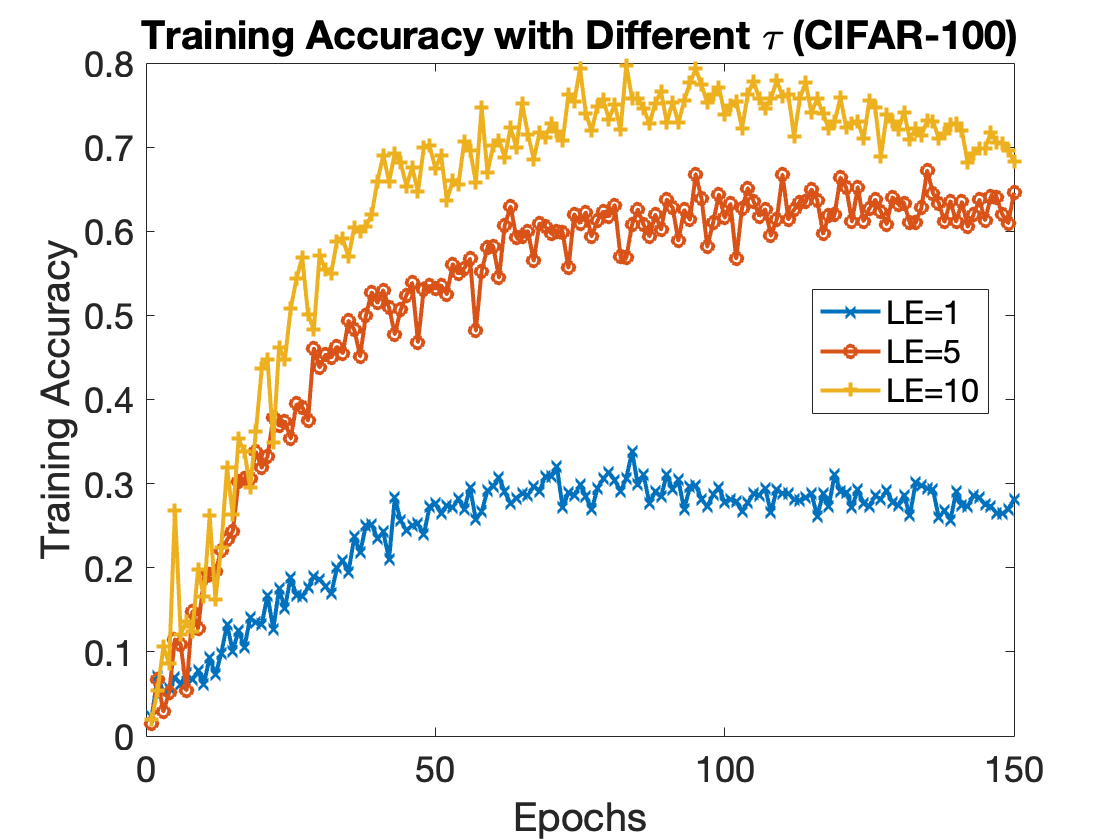}
    \caption{\sl Training accuracy on CIFAR-100.}
    \label{fig:TA_C100}
\end{subfigure}
\caption{\sl \textbf{\algname} training accuracy with different numbers of local steps $\tau$. When the data become more complicated, increasing local epochs may be a better choice.}
\label{fig:TA_LE}
\end{figure*}

We conducted experiments for 150 epochs on the MNIST, CIFAR-10, and CIFAR-100 datasets. As shown in Figure \ref{fig:TA_LE}, increasing the number of local epochs in \textbf{\algname} leads to faster convergence. For \(\tau=1\), the training did not converge even after 150 epochs on MNIST, and for CIFAR-10 and CIFAR-100, it seemed to converge to a lower training accuracy. We observed that as the dataset and model complexity increased, increasing the number of local epochs tended to improve performance. 

Table \ref{tab:tau} presents the final \textbf{\algname} testing accuracies for different numbers of local epochs across the datasets. On MNIST, the testing accuracies were 93.27\% and 93.47\%, respectively, showing only a slight difference, likely because the MNIST dataset is relatively simple, so the learning hyperparameters do not make much of a difference in model performance. For CIFAR-10, the testing accuracies for \(\tau=5\) and \(\tau=10\) were 70.61\% and 66.52\%, respectively, where a larger number of local epochs actually reduced the final performance. However, for CIFAR-100, \(\tau=10\) resulted in the best performance. This suggests that for more complex datasets, a higher number of local epochs can be beneficial, as indicated by the training accuracy curves. Nevertheless, it is important to note that setting \(\tau\) too high may lead to overfitting to the local data, as was the case with \(\tau=10\) on the CIFAR-10 dataset. \new{These findings are consistent with known results in general federated learning, where a higher number of local epochs can effectively increase the number of gradient steps taken, accelerating convergence as long as the local models do not diverge too much due to a large number of local steps.}

\begin{table}[h!]
  \centering
\begin{tabular}
{|p{3.0cm}|p{1.2cm}|p{1.2cm}|p{1.2cm}|}
\hline  Local Epochs & 1 & 5 & 10 \\
\hline MNIST & 74.20 & 93.27 & 93.47 \\
\hline CIFAR-10 & 41.34 & 70.61 & 66.52 \\
\hline CIFAR100 & 19.86 & 43.35 & 44.99 \\
\hline
\end{tabular}
\caption{\sl Final \textbf{\algname} testing accuracies for different number of local epochs.}
  \label{tab:tau}
\end{table}

\subsubsection{Influence of the Final Phase} \label{sec:fl_final}
Our \textbf{\algname}~algorithm uses a final phase that follows the typical federated learning training process. The optimal number of epochs for this final phase varies depending on the dataset and learning model. Due to the simplicity of EMNIST and its model, the testing accuracy is already sufficiently high after aggregation. In our EMNIST setup, using 10 epochs in the final phase increases performance by 0.5\%, and beyond 10 epochs, the testing accuracy stabilizes. For CIFAR-10 and CIFAR-100, the testing accuracy improves by 7\% and 6\%, respectively, after 15 epochs. Around 30 epochs are sufficient to achieve optimal performance for both datasets. It is important to note that choosing the correct number of epochs and learning rate for this final phase is crucial. Too many epochs, or a learning rate that is too high (or with insufficient decay), may lead to overfitting to the local data. Since this final phase is trained locally without any communication overhead, it presents a key advantage of our \textbf{\algname}~algorithm \new{in communication-constrained settings}. Additionally, note that for EMNIST, CIFAR-10, and CIFAR-100, our \textbf{\algname}~already achieves higher accuracies compared to other methods, even without this final phase. Other algorithms like \textbf{FedEM} perform aggregation during the regular training phase, so adding extra local rounds in a final phase of training may lead to overfitting.

\begin{figure*}[h!]
\begin{subfigure}{0.32\textwidth}
    \centering
    \includegraphics[width=1.0\linewidth]{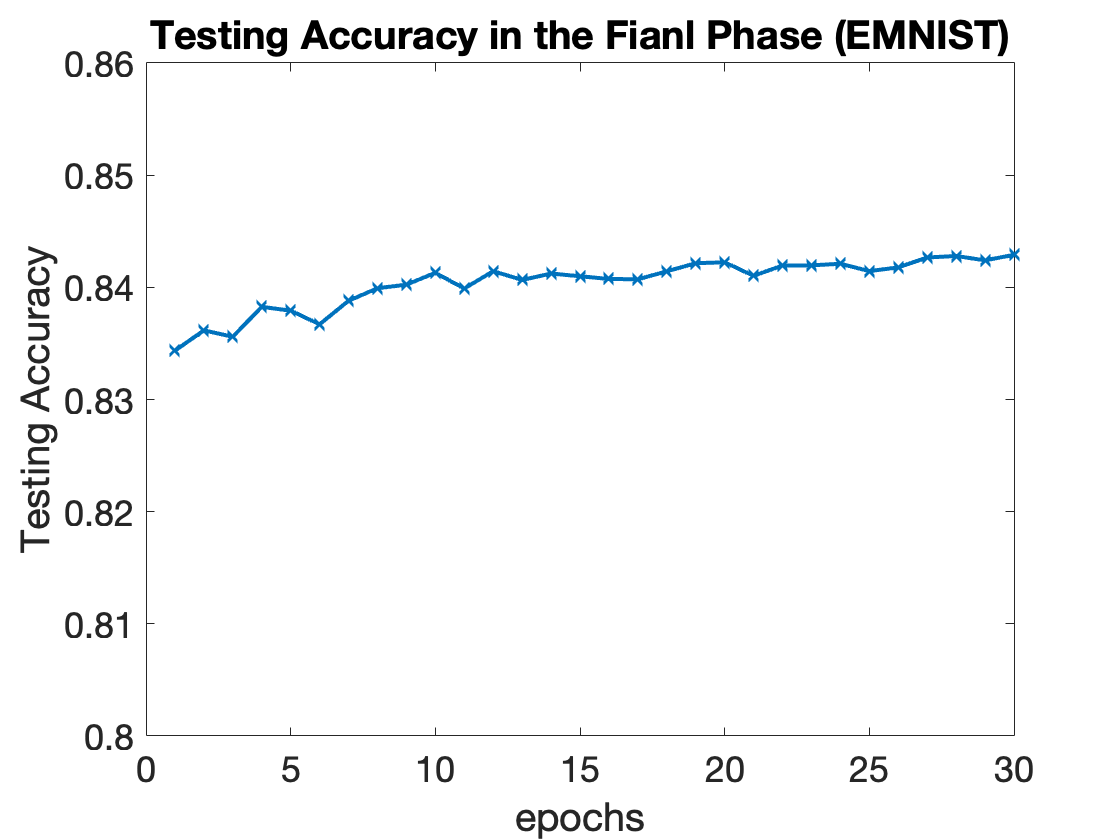}
    \caption{\sl Testing accuracy on EMNIST.}
    \label{fig:TT_EM}
\end{subfigure}
\hfill
\begin{subfigure}{0.32\textwidth}
    \centering
    \includegraphics[width=1.0\linewidth]{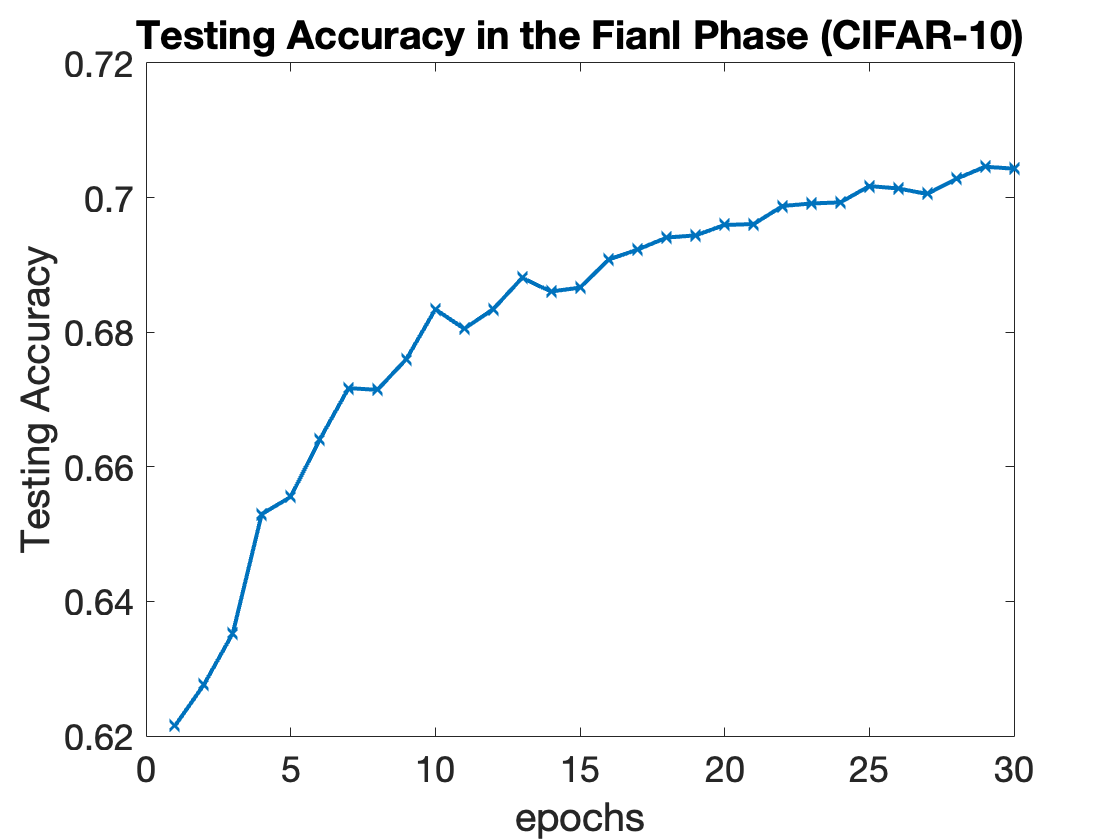}
    \caption{\sl Testing accuracy on CIFAR-10.}
    \label{fig:TT_C10}
\end{subfigure}
\hfill
\begin{subfigure}{0.32\textwidth}
    \centering
    \includegraphics[width=1.0\linewidth]{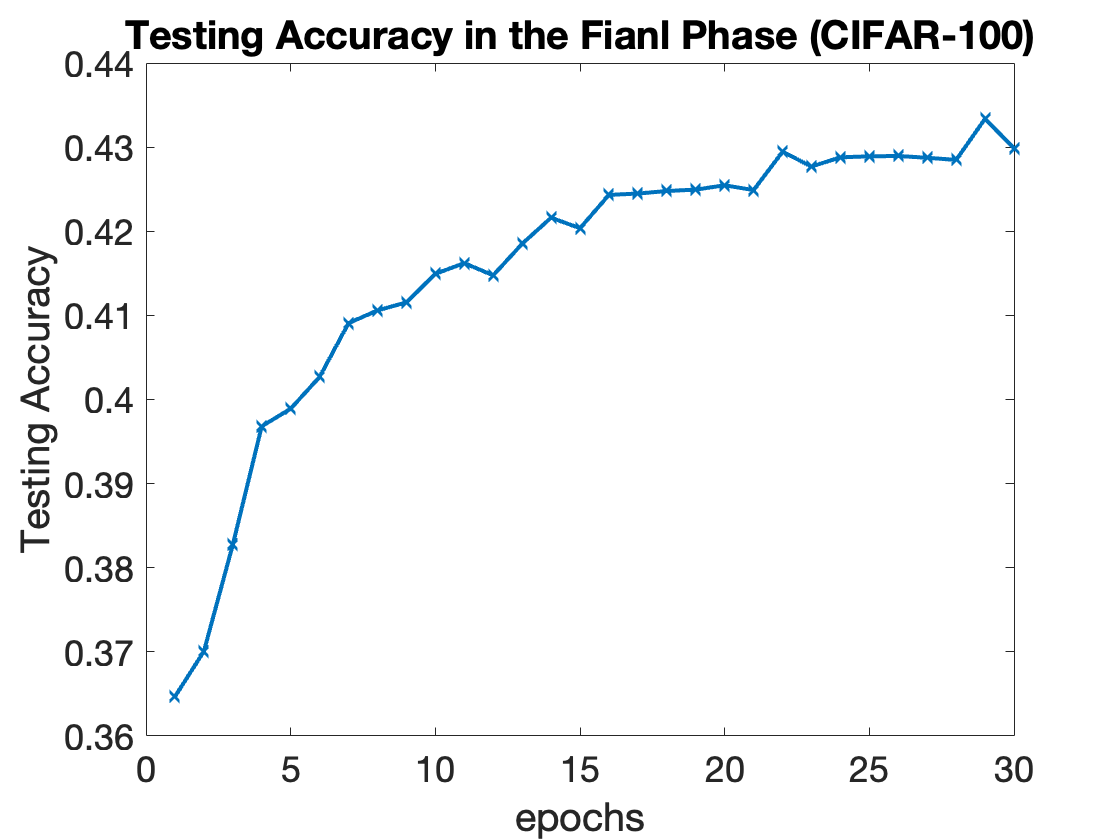}
    \caption{\sl Testing accuracy on CIFAR-100.}
    \label{fig:TT_C100}
\end{subfigure}
\caption{\sl Testing accuracy of the final phase.}
\label{fig:TT_FP}
\end{figure*}

\subsubsection{Influence of the Number of Clusters} \label{sec:fl_cluster}
\begin{figure}
    \centering
    \includegraphics[width=0.5\linewidth]{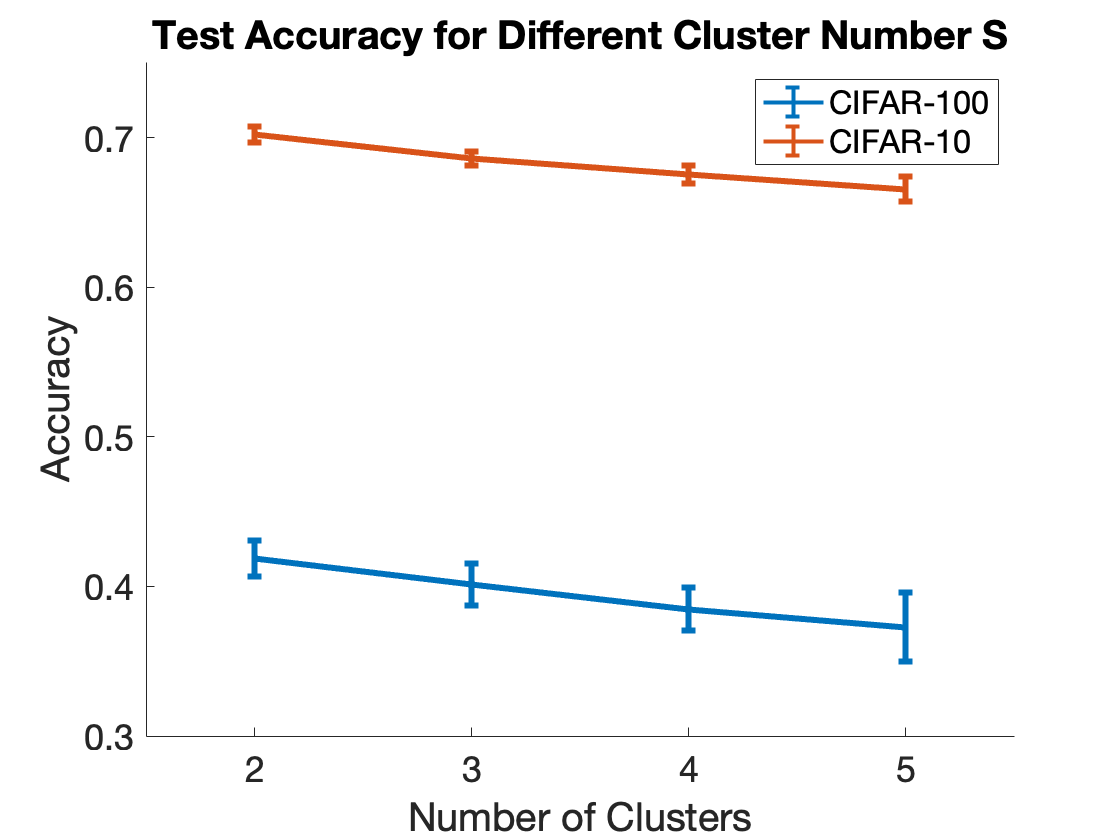}
    \caption{\textbf{\algname} test accuracy for different numbers of clusters $S$.}
    \label{fig:cns}
\end{figure}

The testing accuracy with different hyperparameters $S$ (number of clusters) for the CIFAR-10 and CIFAR-100 datasets is shown in Figure \ref{fig:cns}. In the experimental settings, we potentially created four different distributions by using varying labels and image rotations. In our \textbf{\algname}~algorithm, setting $S$ too high does not necessarily improve performance. This may be because most practical loss functions, such as the cross-entropy used in neural networks, are non-convex, meaning that the aggregated model may not perform optimally in practice. Aggregating more models in the final phase can exacerbate this issue. However, in our \textbf{\algname}~algorithm, setting $S=2$ already gives excellent performance in terms of the final test accuracy.

\subsubsection{Extra Details for Experiments with Different Graph Connectivity} \label{sec:fl_edcon}
\begin{figure*}[h!]
\begin{subfigure}{0.32\textwidth}
    \centering
    \includegraphics[width=1.0\linewidth]{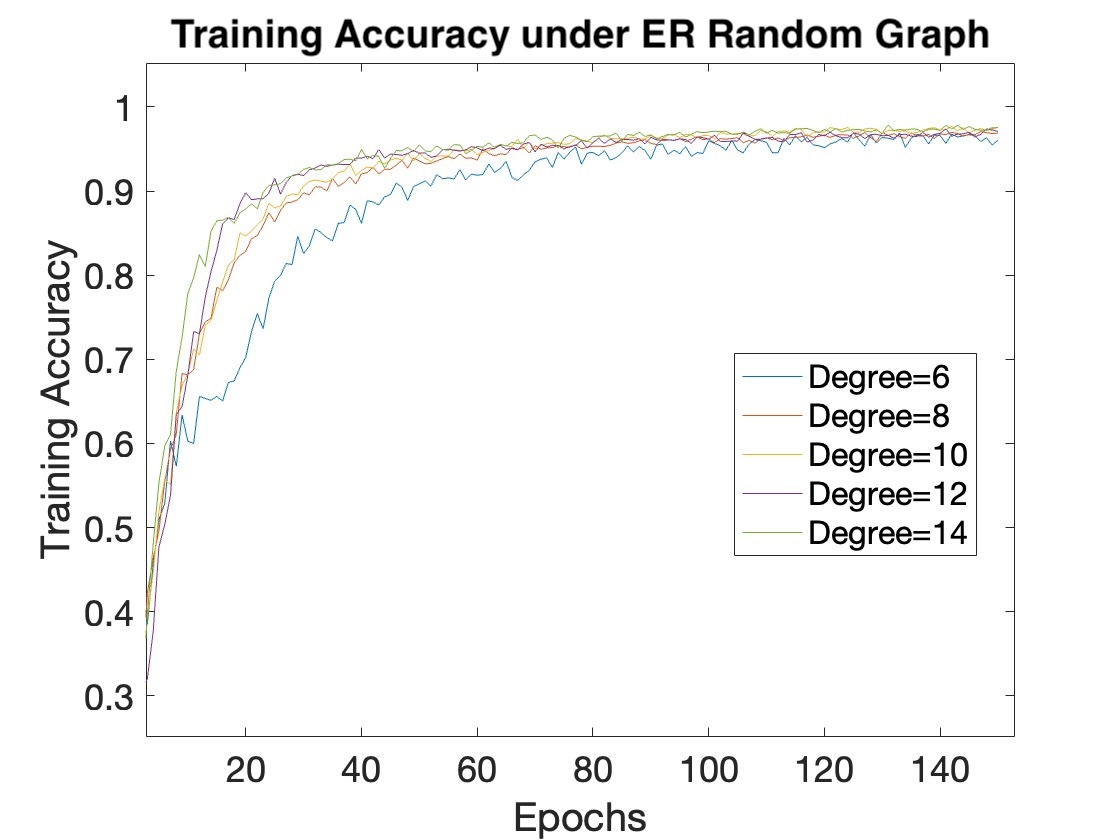}
    \caption{\sl Training accuracy of ER Graph.}
    \label{fig:ER}
\end{subfigure}
\hfill
\begin{subfigure}{0.32\textwidth}
    \centering
    \includegraphics[width=1.0\linewidth]{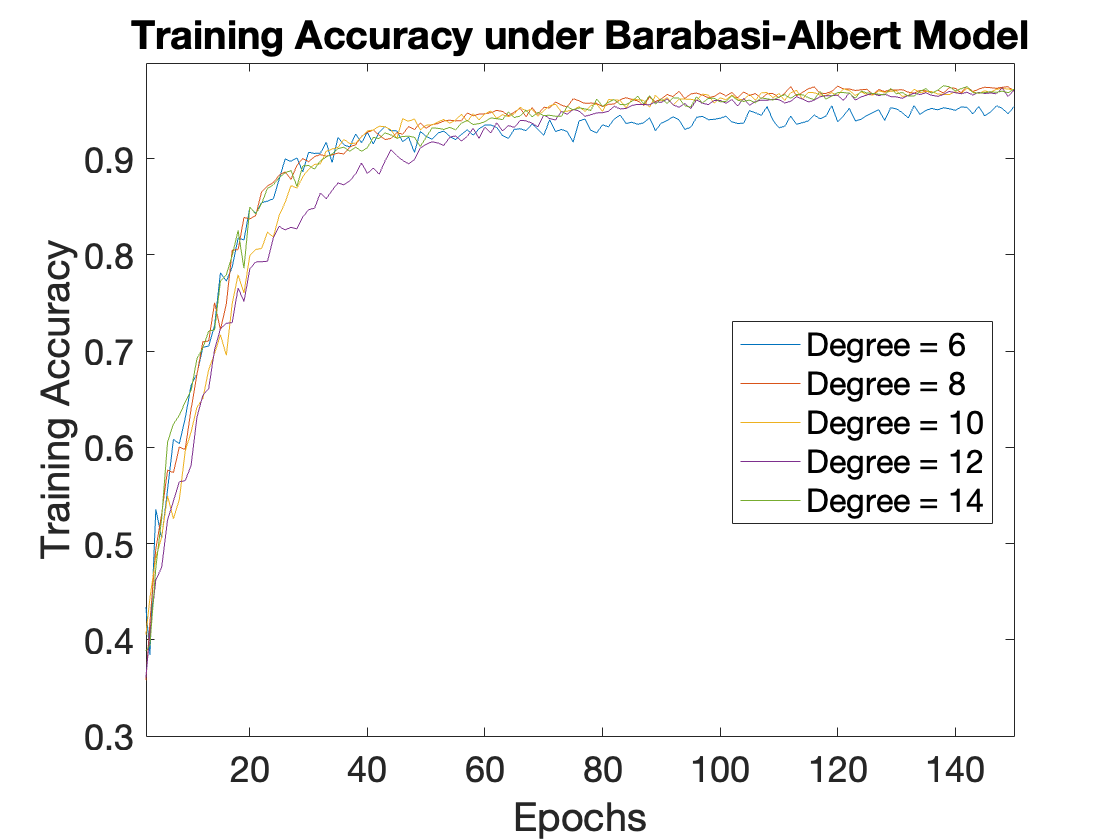}
    \caption{\sl Training accuracy of BA Model.}
    \label{fig:BA}
\end{subfigure}
\hfill
\begin{subfigure}{0.32\textwidth}
    \centering
    \includegraphics[width=1.0\linewidth]{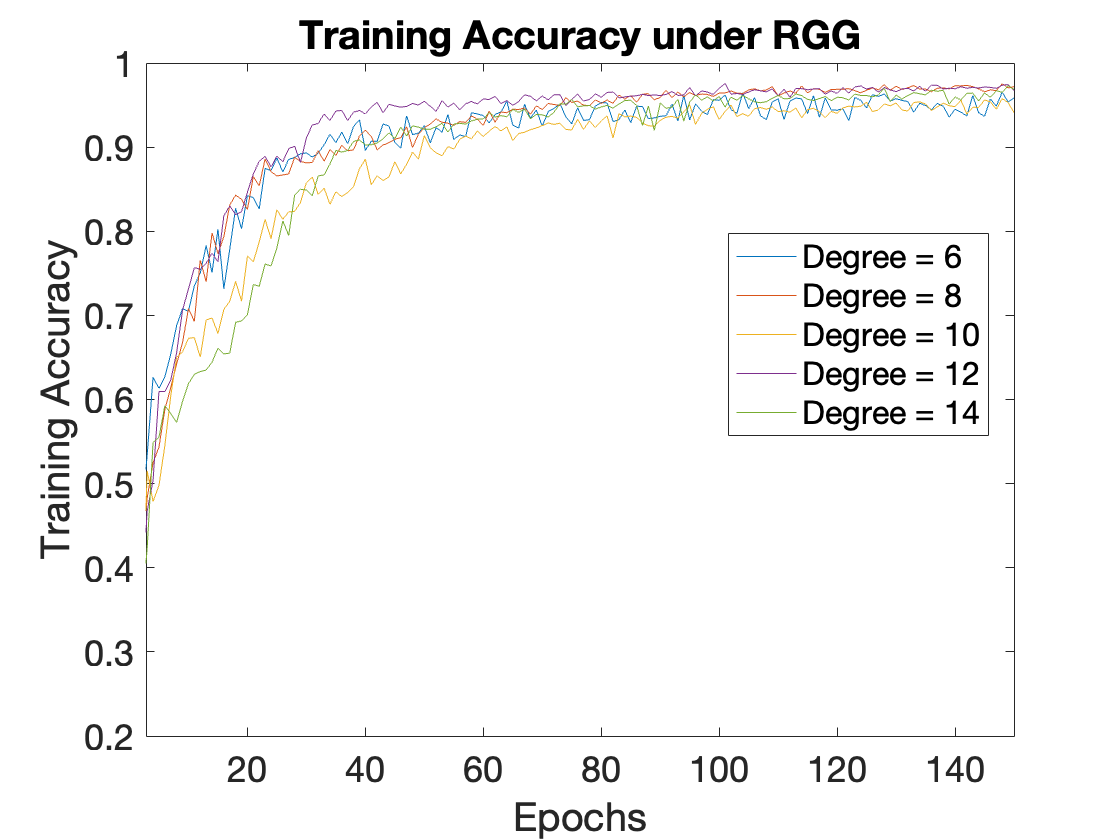}
    \caption{\sl TrainingaAccuracy of RGG.}
    \label{fig:RGG}
\end{subfigure}
\caption{\sl \textbf{\algname}~converges slightly faster on networks of higher average degree, with noisier convergence on highly clustered RGG graphs, on MNIST Data.}
\label{fig:topology}
\end{figure*}

\textbf{\algname}'s training accuracy versus epochs for MNIST across different topologies is shown in Figure \ref{fig:topology}. We observe that networks with lower connectivity typically converge more slowly than those with higher connectivity, in each topology. Additionally, RGG exhibits more oscillations compared to other topologies, likely due to its high clustering effect \citep{penrose2003random}. However, all topologies eventually reach the same level of training accuracy, regardless of the network structure, indicating that, as predicted by Theorem~\ref{thm:4}, \textbf{\algname}~converges as long as the network is connected.

In addition to static network topology settings, we evaluate the performance of our \textbf{\algname} algorithm under dynamic network conditions. In the following experiments, we use CIFAR-100 with 25 clients, initializing the network with an Erdős--Rényi (ER) random graph. To simulate a dynamic network, at each epoch, every existing edge has a probability $p$ of being removed, while each non-existent edge has a probability $p_{\text{add}}$ of being added. The value of $p_{\text{add}}$ is adjusted at each epoch to maintain a roughly constant average connectivity across the network. A larger value of $p$ corresponds to a more dynamic network topology over time. The results are summarized in Table \ref{tab:dynet}. From the results, we observe that network dynamics have little effect on performance---our \textbf{\algname} consistently maintains its effectiveness across different edge removal probabilities $p$.

\begin{table}[h!]
  \centering
\begin{tabular}
{|p{2.4cm}|p{1.2cm}|p{1.2cm}|p{1.2cm}|p{1.6cm}|}
\hline  $p$ & 0.3 & 0.2 & 0.1 & 0 (Static) \\
\hline Test Accuracy & 37.56 & 37.22 & 37.42 & 37.14 \\
\hline
\end{tabular}
\caption{\sl Performance of Dynamic Network Topology.}
  \label{tab:dynet}
\end{table}

\subsubsection{Impact of Data Quantity Imbalance Across Clients}\label{sec:IDA}

In addition to considering data imbalance across clusters, we further evaluate the performance of our method under a more challenging setting where both inter-cluster imbalance and total data imbalance across clients are present. Specifically, we conduct experiments using the CIFAR-100 dataset with the same configuration as described in Figure~\ref{fig:CFCN}. To simulate varying amounts of data per client, we categorize clients into three groups: low, average, and high data holders. Let $r$ denote the ratio of data volume between clients with the highest and lowest data quantities.

\begin{figure*}[h!]
\centering
\begin{subfigure}{0.49\textwidth}
    \centering
    \includegraphics[width=1.0\linewidth]{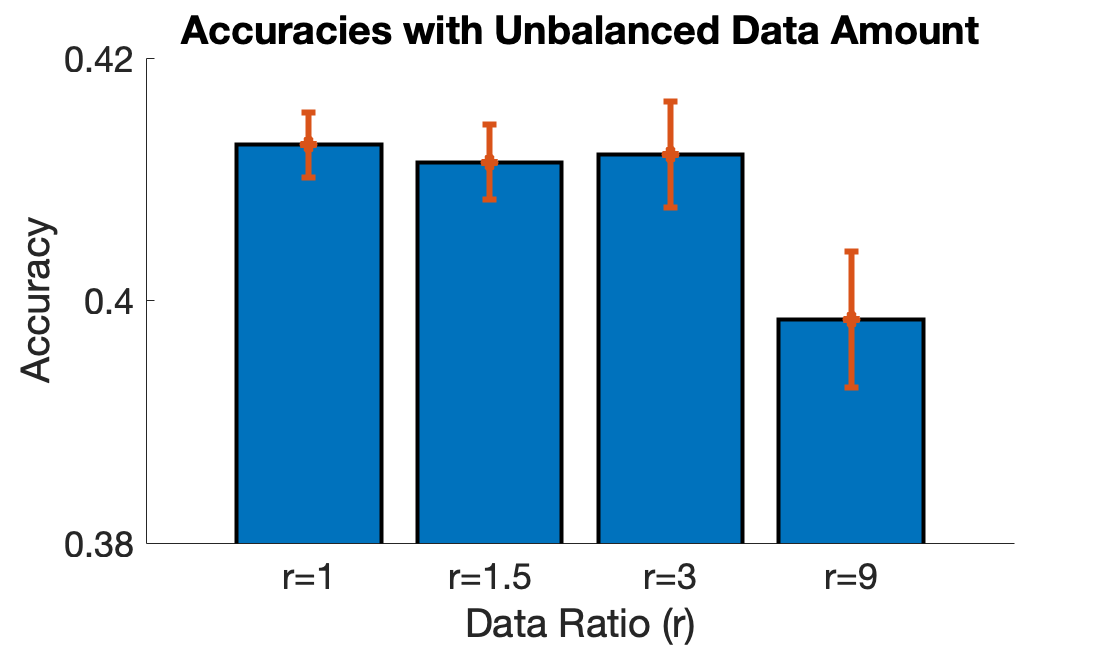}
    \caption{\sl Average Test accuracy with unbalanced data amount.}
    \label{fig:UDNA}
\end{subfigure}
\begin{subfigure}{0.49\textwidth}
    \centering
    \includegraphics[width=1.0\linewidth]{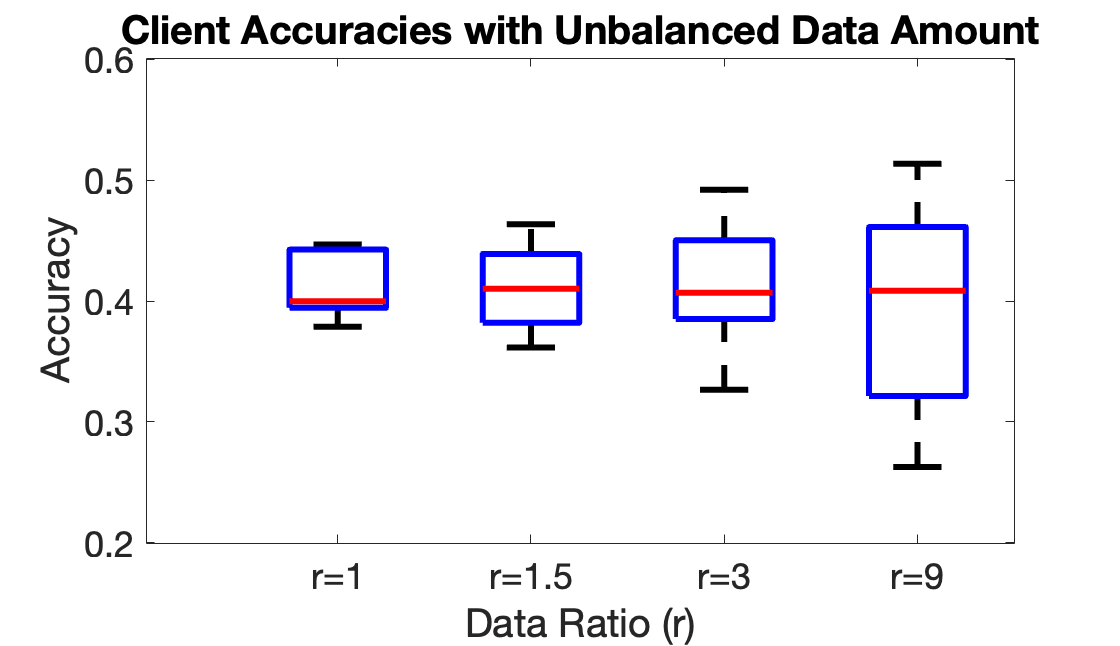}
    \caption{\sl Box plot of test accuracy across clients.}
    \label{fig:UDNB}
\end{subfigure}
\caption{\sl Test accuracy with unbalanced data amount.}
\label{fig:UDN}
\end{figure*}

The results of this experiment are presented in Figure~\ref{fig:UDN}. We observe that the average accuracy remains stable as the imbalance ratio $r$ increases. Notably, even under the most skewed setting ($r=9$), the clients with the lowest test accuracy achieve approximately $30\%$ accuracy—substantially higher than the performance of local training under uniform data allocation, which yields only around $14\%$ accuracy. This demonstrates that clients with limited data can significantly benefit from collaborative training and knowledge sharing with other clients.

\subsubsection{Experiments Incorporating Differential Privacy (DP)}\label{sec:dp}

We follow \cite{wei2020federated} conduct the experiments on MNIST dataset with 50 clients. The parameters of DP is selected as follow: Clipping Threshold $C = 1$, $\delta = 0.01$ thus $c$ is chosen to be $\sqrt{2 \ln \frac{1.25}{0.01}}$. We select $\epsilon$ to be 10, 50 and 100 to do the experiment. Table \ref{tab:dp} shows the results of different settings with two different accuracies. One is the test accuracy of our \textbf{\algname} right after the model aggregation. The other one is the accuracy after 10 local epochs of our final phase. The reason to include 2 different accuracies is because the final phase is local training, need not to do the DP. If we only show the accuracy after the final phase, the influence of DP might not be clear.

\begin{table}[h!]
  \centering
\begin{tabular}
{|p{5.0cm}|p{2cm}|p{2cm}|p{2cm}|p{2cm}|}
\hline  Metrics & No DP & DP ($\epsilon=100$) & DP ($\epsilon=50$) & DP ($\epsilon=10$) \\
\hline Test Accuracy (Post Aggregation) & 92.51 & 92.75 & 92.46 & 92.13 \\
\hline Test Accuracy (After Final Phase) & 93.89 & 93.99 & 93.87 & 93.70 \\
\hline
\end{tabular}
\caption{\sl Results with DP on MNIST dataset.}
  \label{tab:dp}
\end{table}

From the results, we see that our \textbf{\algname} algorithms combine perfectly with DP. The accuracies keep at a high level with different settings. The $\epsilon=100$ case even have a slightly higher accuracy compare to the case without DP. This may be potentially due to a moderate additive noise actually preventing over-fitting in certain level. Another observation is that, actually the final phase of our algorithms do enhance the model and reduce the gap of the test accuracies across different settings. This is another evidence showing that the final phase of our \textbf{\algname} do further personalize the local model well.

\subsubsection{Experiments Using MobileNet-v2}\label{sec:mobilenet}
We conducted experiments on the CIFAR-100 dataset and the CIFAR-10 dataset and its mixtures with MNIST and FashionMNIST, using MobileNet-v2 as the machine learning model. The mixed datasets were created by sampling 25,000 data points from CIFAR-10 and 25,000 from MNIST/FashionMNIST. Each client randomly drew between 10\% and 90\% of its data from one of the sampled datasets, with the remainder sourced from the other. The network topology was modeled as an Erdős–Rényi (ER) random graph with a connection probability of $p = 0.20$ and a total of 20 clients. The results are presented in Table~\ref{tab:mb}.

For CIFAR-100, our method outperforms other methods. However in the data mixture settings, the results indicate that as the model size and complexity increase, FedAvg outperforms all other algorithms. As models and datasets become more complex, personalization methods may occasionally exhibit reduced performance due to many reasons. This may be attributed to the model’s expressiveness, which allows it to effectively capture variations across different clients. In this case, the global model is sufficiently robust and more effective than personalizing to local data distributions. Similar results were also observed in the decentralized FedEM \cite{marfoq2021federated}, where performance degradation occurred under specific conditions in their experiments. In specific, their decentralized FedEM performs worse than FedAvg on FEMNIST and CIFAR-10.

\begin{figure*}[h!]
\begin{subfigure}{0.32\textwidth}
    \centering
    \includegraphics[width=1.0\linewidth]{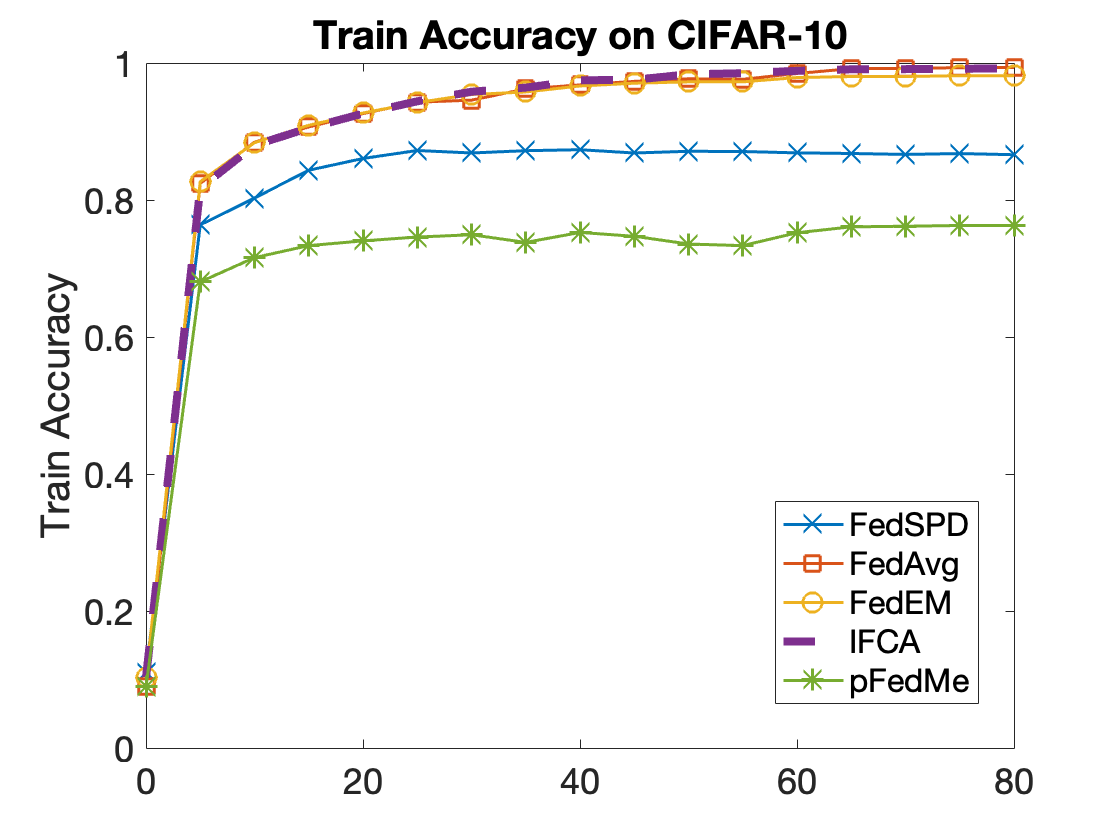}
    \caption{\sl Training Accuracy of Single CIFAR-10 Dataset.}
    \label{fig:mb-cf}
\end{subfigure}
\hfill
\begin{subfigure}{0.32\textwidth}
    \centering
    \includegraphics[width=1.0\linewidth]{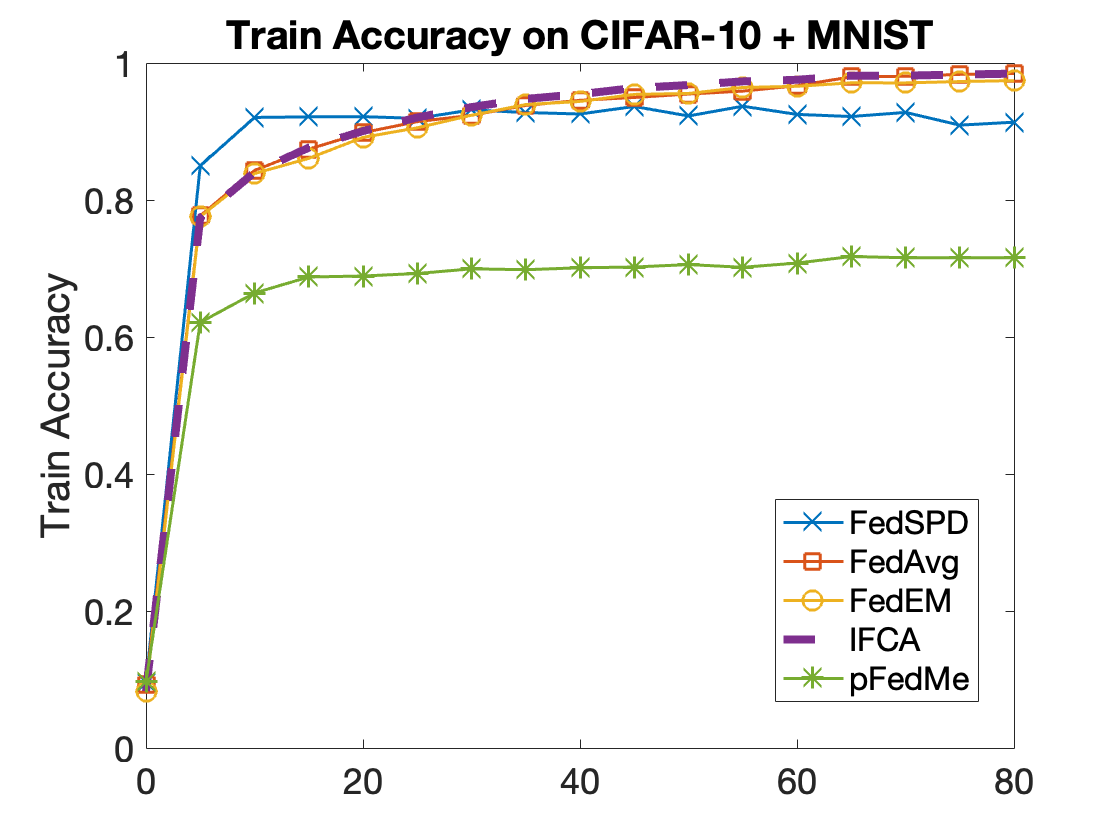}
    \caption{\sl Training Accuracy of Mixture of CIFAR-10 + MNIST.}
    \label{fig:mb-mm}
\end{subfigure}
\hfill
\begin{subfigure}{0.32\textwidth}
    \centering
    \includegraphics[width=1.0\linewidth]{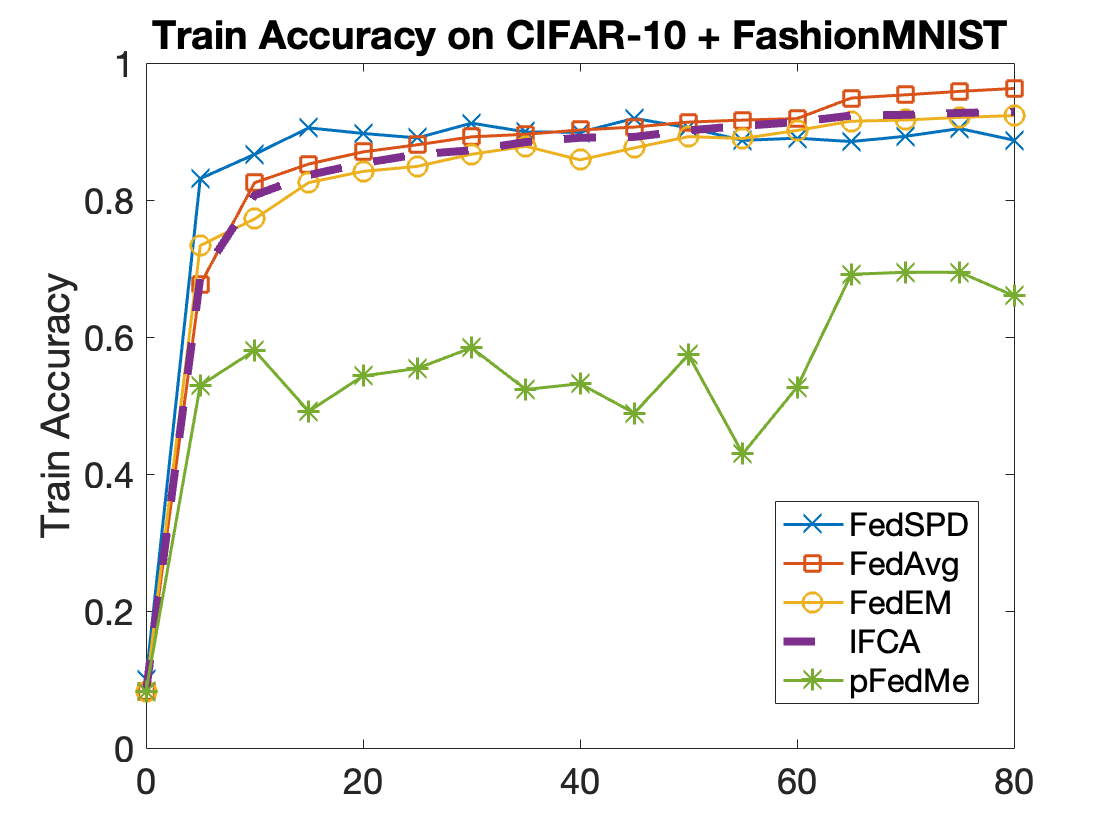}
    \caption{\sl Training Accuracy of Mixture of CIFAR-10 + FashionMNIST.}
    \label{fig:mb-mf}
\end{subfigure}
\caption{\sl Experiments on Various Datasets Using MobileNet-v2.}
\label{fig:mobile}
\end{figure*}

Our proposed method, \textbf{\algname}, experiences greater challenges in such scenario, as it splits the local dataset into two clusters for separate training and postpone the aggregation. However, when the distributions of the two clusters differ significantly, such as in the mixture of CIFAR-10 with MNIST or FashionMNIST, \textbf{\algname} achieves faster convergence, as demonstrated in Figures~\ref{fig:mb-mm} and~\ref{fig:mb-mf}. This highlights \textbf{\algname}'s ability to accurately distinguish data sampled from different distributions, showing that \textbf{\algname} can be trained efficiently when the communication/computing resources are limited.

\begin{table}[htbp]
  \centering
\begin{tabular}
{|p{4.5cm}|p{1.5cm}|p{1.5cm}|p{1.5cm}|p{1.5cm}|p{1.5cm}|p{1.5cm}|}
\hline   & \textbf{FedSPD} & FedEM & IFCA & pFedMe & FedAvg & Local \\
\hline CIFAR-10 & 72.14 & 78.02 & 78.56 & 61.00 & 79.01 & 57.72\\
\hline CIFAR-10 + MNIST & 86.20 & 88.88 & 89.08 & 74.64 & 89.27 & 77.76\\
\hline CIFAR-10 + FashionMNIST & 80.04 & 85.01 & 85.47 & 69.37 & 85.58 & 72.74\\
\hline CIFAR-100 & 46.13 & 42.29 & 44.48 & 32.77 & 45.70 & 18.52\\
\hline
\end{tabular}
\caption{\sl Results using MobileNet-v2.}
  \label{tab:mb}
\end{table}

%% file: sections/compare.tex
\section{Detailed comparison between \textbf{\algname} and \textbf{FedSoft}}\label{sec:comp}

The training processes of \textbf{\algname} and \textbf{FedSoft} differ fundamentally. In \textbf{FedSoft}, local training utilizes a proximal objective function with regularization terms to account for the distance to each cluster center. Conversely, \textbf{\algname} trains using data associated with the selected cluster, where model parameters are updated based on an objective function that does not include the regularization term. This modification is critical in decentralized federated learning, where ensuring both convergence to an optimal solution and consensus on cluster centers is essential. Unlike centralized systems, which aggregate cluster centers at a single server, decentralized systems lack this central coordination, making consensus challenging.
\textbf{\algname} was inspired by \textbf{FedSoft}. Attempts were made to bound the consensus distance of cluster centers in \textbf{FedSoft}. However, the results suggest that in decentralized settings of \textbf{FedSoft}, the consensus may not be reached. Experimental results in Section 6 demonstrate \textbf{FedSoft}'s limitations in the decentralized scenarios. Specifically, \textbf{FedSoft}'s cluster centers fail to reach the optimal values due to its update rules:

\begin{itemize}
    \item Uniform Data Utilization: \textbf{FedSoft} updates each cluster center using all available data.
    \item Probabilistic Contribution: \textbf{FedSoft} uses probabilities proportional to the estimated data distribution among clusters to guide contributions to the selected cluster.
\end{itemize}

These update rules lead to gradients during local updates being biased towards a mixture of optimal cluster centers from all clusters, rather than the correct optimal center for the selected cluster. As datasets grow more complex and the optimal cluster centers diverge significantly (e.g., CIFAR-10 or CIFAR-100), this bias becomes more pronounced, causing degraded performance. This is evident in Section 6, where \textbf{FedSoft}'s performance deteriorates as the datasets shift from EMNIST to CIFAR-10 and CIFAR-100 in both centralized and decentralized scenarios.

In decentralized settings, the sparse updates exacerbate the difficulty for clients to estimate optimal cluster centers accurately. This is especially problematic for complex datasets like CIFAR-10 and CIFAR-100, where \textbf{FedSoft} performs significantly worse. This limitation highlights the need for a different approach, such as the one introduced by \textbf{\algname}.

In sum, in \textbf{\algname} during each round of the first step, clients train separate models using data associated with their selected clusters. This approach ensures consensus and optimality of the cluster centers. Unlike centralized scenarios, where clients share a unified cluster center, each client in decentralized systems maintains different cluster centers. This necessitates rigorous proof of consensus across clients, as described in our theoretical analysis. Thus, the proof techniques of \textbf{\algname} and \textbf{FedSoft} are completely different.

Once consensus on cluster centers is achieved, the second phase of \textbf{\algname} aggregates models by computing a weighted average of the cluster centers, aligning with \textbf{FedSoft}’s objective. To address the suboptimality of non-convex models, \textbf{\algname} incorporates an additional final phase of local training. This phase enables further exploration of the model parameters, mitigating suboptimality and enhancing overall performance.

\textbf{Differences in Theoretical Analysis.}
\begin{itemize}
    \item Relax of Assumption 2 in the \textbf{FedSoft}. Since \textbf{FedSoft} requires the Assumption 2 in their paper which state the $\beta$ similarity among all subproblems of different clusters. In other words, the optimals of different cluster centers need to be close enough to guarantee the bounded distance between the learned cluster center and the optimal cluster center. This is because they use all data to update the cluster center. If using the data from other cluster to update the selected cluster, this assumption is required to guarantee the gradient update is not going to far away from the optimal of the selected cluster. In contrast, the different update rule of our algorithm \textbf{\algname} on cluster center further guarantee the optimality of our algorithms and have a tighter bound of convergence without the requirement of this additional assumption.
    \item The proof of \textbf{FedSoft} is based on the centralized FL (CFL). Thus, consensus is automatically met with the centralized aggregation. However, they do not proof the effectiveness under the decentralized settings. In contrast, we proof the consensus of the cluster centers in \textbf{\algname} under decentralized FL. This is one of the main challenges of the theoretical analysis. Proving the consensus and convergence is much more difficult in decentralized FL (DFL) than the CFL, since all clients keep the different estimation of the cluster centers. Our attempts try to bound the consensus distance of \textbf{FedSoft} in DFL was failed since the way that FedSoft update its cluster center may not yield the same optimal for all clients. It may differ based on different neighboring clients.
\end{itemize}